\documentclass[10pt,twocolumn,letterpaper]{article}

\usepackage[pagenumbers]{cvpr} 

\usepackage{algorithm}
\usepackage{algorithmic}
\usepackage{amsfonts}
\usepackage{amsmath}
\usepackage{amssymb}
\usepackage{arydshln}
\usepackage{bbm}
\usepackage{bm}
\usepackage{booktabs}
\usepackage{dashrule}
\usepackage{graphicx}
\usepackage{makecell}
\usepackage{diagbox}
\usepackage{multirow}
\usepackage{pifont} 

\newcommand{\xmark}{\ding{55}}
\usepackage{subcaption}
\usepackage{tcolorbox}
\usepackage{wrapfig}
\usepackage{url}
\usepackage[table]{xcolor}         
\usepackage{mathtools}

\definecolor{darkred}{rgb}{0.6, 0.16, 0.25}

\definecolor{cornellred}{rgb}{0.7, 0.11, 0.11}
\definecolor{cadmiumgreen}{rgb}{0.0, 0.42, 0.24}
\definecolor{Blue9}{rgb}{0.098,0.3,0.9}

\definecolor{backred}{RGB}{255, 190, 190}
\definecolor{backblue}{RGB}{210, 230, 250}

\definecolor{backblue2}{RGB}{212, 221, 239}
\definecolor{Grayheavy}{gray}{0.90}

\definecolor{demphcolor}{RGB}{144,144,144}

\definecolor{cvprblue}{rgb}{0.21,0.49,0.74}
\usepackage[pagebackref,breaklinks,colorlinks,allcolors=cvprblue]{hyperref}

\title{Unified Generation and Self-Verification for Vision-Language Models \\via Advantage Decoupled Preference Optimization}

\author{
Xinyu Qiu$^{1,2}$\thanks{Equal contribution.}\quad
Heng Jia$^{1,2}$\footnotemark[1]\quad
Zhengwen Zeng$^{2}$\quad
Shuheng Shen$^{2}$\thanks{Corresponding Authors.}\quad
Changhua Meng$^{2}$\quad
Yi Yang$^{1}$\quad
Linchao Zhu$^{1}$\footnotemark[2]
\\
$^{1}$College of Computer Science and Technology, Zhejiang University \\
$^{2}$Venus Team, Ant Group \\
{\tt\small \{qiuxy020,jiaheng.dlut,zhulinchao7\}@gmail.com} \\
{\tt\small \{zengzhengwen.zzw,shuheng.ssh,changhua.mch\}@antgroup.com} \\
}

\begin{document}
\maketitle

\begin{abstract}
    Parallel test-time scaling typically trains separate generation and verification models, incurring high training and inference costs.
    We propose Advantage Decoupled Preference Optimization (\textbf{ADPO}), a unified reinforcement learning framework that jointly learns answer generation and self-verification within a single policy.
    ADPO introduces two innovations: \textbf{a preference verification reward} improving verification capability and \textbf{a decoupled optimization mechanism} enabling synergistic optimization of generation and verification.
    Specifically, the preference verification reward computes mean verification scores from positive and negative samples as decision thresholds, providing positive feedback when prediction correctness aligns with answer correctness.
    Meanwhile, the advantage decoupled optimization computes separate advantages for generation and verification, applies token masks to isolate gradients, and combines masked GRPO objectives, preserving generation quality while calibrating verification scores.
    ADPO achieves up to \textbf{+34.1\%} higher verification AUC and \textbf{-53.5\%} lower inference time, with significant gains of \textbf{+2.8\%/+1.4\%} accuracy on MathVista/MMMU, \textbf{+1.9} cIoU on ReasonSeg, and \textbf{+1.7\%/+1.0\%} step success rate on AndroidControl/GUI Odyssey.
\end{abstract}

\begin{figure*}[htbp]
    \centering
    \includegraphics[width=\linewidth]{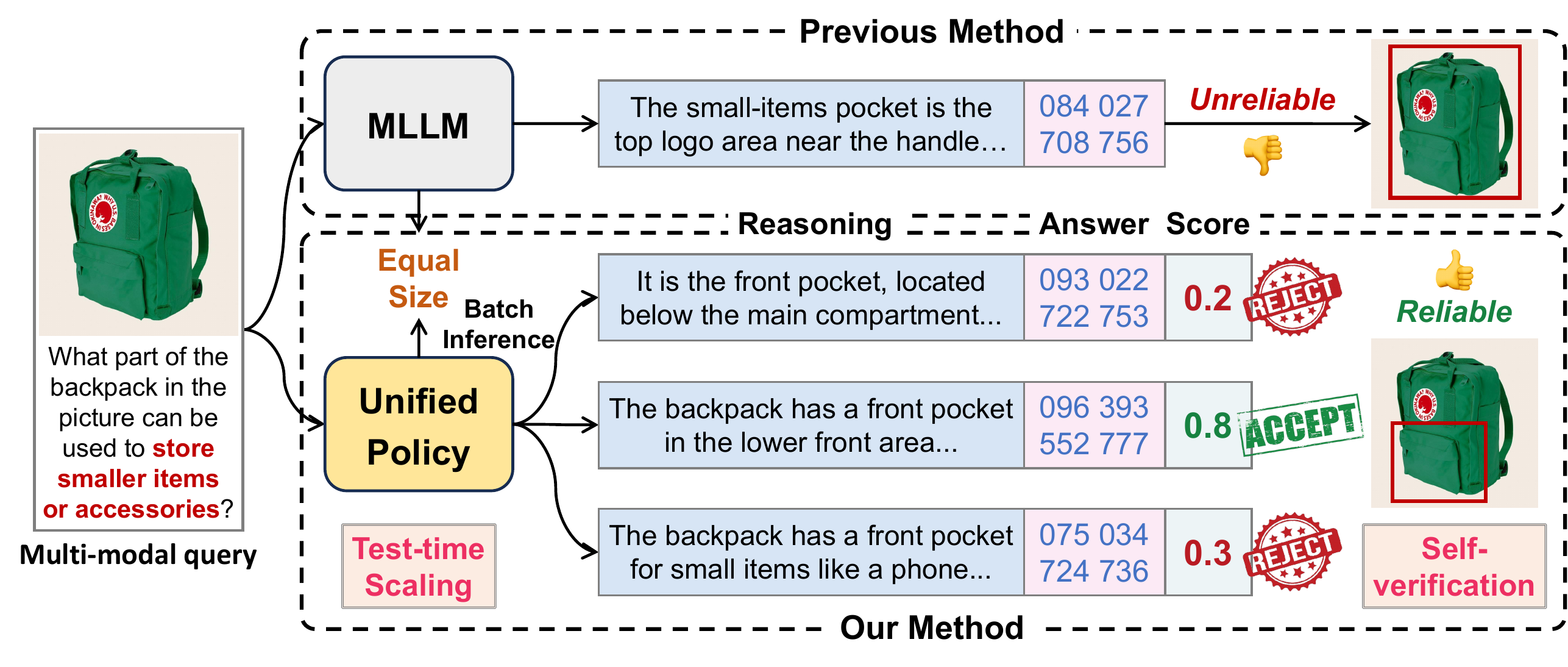}
    \caption{
        \textbf{Overview of ADPO.}
        We build a unified reinforcement learning framework that jointly learns answer generation and self-verification within a single policy.
        The unified policy model provides reliable scoring that enables effective test-time scaling via best-of-N selection, significantly improving performance across multimodal tasks.
    }
    \label{fig:00_teaser}
\end{figure*}

\section{Introduction}
Test-time scaling enhances reliability and performance by spending more compute at inference time, which can generally be realized in two ways~\cite{o1,r1,muennighoff2025s1}.
Serial test-time scaling enables LLMs to enhance problem-solving capabilities by sequentially generating additional thinking tokens~\cite{muennighoff2025s1,wang2025faster}.
Parallel test-time scaling generates multiple solutions simultaneously, then aggregates candidates and selects the best one~\cite{self-consistency,yao2023tree}.

Serial test-time scaling, as demonstrated by DeepSeek-R1~\cite{r1} and OpenAI-o1~\cite{o1}, achieves substantially improved performance in mathematics and coding domains.
However, when transferring to multimodal domains, recent works have found that serial test-time scaling provides only limited performance improvements on image classification, video understanding and visual spatial understanding tasks~\cite{li2025think,liao2025improved,lin2025investigating,zeng2025revisiting,tang2025gui}.
These observations highlight the inherent limitations of serial test-time scaling and critically motivate the development of principles for parallel test-time scaling that more effectively support robust and efficient multimodal reasoning~\cite{you2025parallel,chu2025ssr,wen2025parathinker}.

Current approaches ~\cite{sun2025mm,mm-prm} deploy two separate models: a generator that creates potential solutions and a verifier that evaluates and ranks these candidates, leading to substantial performance gains.
However, this methodology suffers from two major limitations: 1) \textit{training resource intensive}, as it requires preparing two specialized datasets and training two independent models; 2) \textit{deployment inefficiency}, since both models must operate concurrently during inference, demanding significant computational resources.
Alternatively, training only a generator (with majority voting for selection) or only a verifier (using a base model for generation) yields limited performance improvements compared to training both components~\cite{sun2025mm, self-consistency,cobbe2021training}.
To address these limitations, we propose a novel reinforcement learning framework that trains a unified policy model to concurrently generate both solutions and self-verification scores of the solutions (\cref{fig:00_teaser}).

We propose Advantage Decoupled Preference Optimization (ADPO), a unified framework that enables reliable self-verification while maintaining generation quality.
This unified paradigm faces two challenges.
First, training the model to verify its own outputs using binary rewards creates severe class imbalance. As the model improves, the proportion of correct answers increases substantially, causing verification scores to collapse toward uniform predictions and eliminating gradient signals (\cref{par:imbalance-challenge}). Second, naively aggregating answer and verification rewards leads to reward hacking. The model exploits the objective by deliberately producing incorrect answers while assigning low verification scores, achieving high total rewards while severely degrading generation performance.

To address the class imbalance challenge, we introduce a preference verification reward that reframes verification as a ranking problem. Instead of comparing scores against fixed thresholds, our approach adaptively partitions samples into positive/negative groups based on answer quality, then rewards the model when verification scores respect the relative quality ordering within each group. By computing advantages from pairwise comparisons rather than absolute labels, this formulation maintains informative gradient signals even under severe class imbalance.

To address the reward hacking challenge, we introduce advantage decoupled optimization that computes separate advantages for answer rewards and preference rewards. Specifically, answer advantages are estimated from answer rewards only, while preference advantages are derived solely from preference rewards. We then apply disjoint token-level masks to isolate gradient flows. In this way, answer advantages exclusively update generation tokens, while preference advantages solely affect verification score tokens. This decoupled computation ensures that improvements in generation quality are driven only by answer-based gradients, while verification calibration is shaped only by preference-based gradients, thereby eliminating reward hacking.
Our contributions are summarized as follows:

\noindent\textbf{1. Preference verification reward.}
We develop Preference Verification Reward, which maintains informativeness under severe class imbalance while improving calibration and robustness.

\noindent\textbf{2. Advantage decoupled optimization.}
We introduce a principled approach to disentangle content generation and verification learning within a unified GRPO framework.

\noindent\textbf{3. Comprehensive evaluation.}
Our method significantly improves task performance and verification quality: ADPO achieves +2.8/+1.4 accuracy gains on MathVista/MMMU, +1.9 cIoU on ReasonSeg, and +1.7/+1.0 step success rates on AndroidControl/GUI Odyssey.

\section{Related Work}

\noindent\textbf{Test-Time Scaling.}
Recent work scales reasoning at test time via longer thinking tokens and majority voting for LLMs \cite{r1,o1,self-consistency,shao2024deepseekmath}.
Multimodal variants adapt this paradigm with R1-style objectives and structured CoT for VLMs \cite{visionreasoner,skywork-r1v,visual-rft,shen2025vlm,vision-r1,zhang2025r1,r1-onevision}.
In agentic settings, GUI agents adopt RL with explicit reasoning traces \cite{ui-r1,infigui-r1,qin2025uitars,huang2025spiritsight,zhang2025agentcpm,ui-venus}.
However, recent ``no-think'' results suggest that more internal tokens do not always translate to better multimodal reasoning \cite{li2025think,liao2025improved}.
We instead couple solution generation with a learned self-verification signal, enabling reliable performance scaling through best-of-$N$ selection without fragile dependence on longer chains.

\noindent\textbf{Multimodal Reward Modeling and Generative Verifiers.}
Another line studies reward modeling for multimodal alignment, including RLHF-style pipelines and chain-of-thought verification \cite{mm-rlhf,sun2025mm}.
Process or scalar reward models provide step-level or outcome supervision for reasoning \cite{mm-prm,dreamprm,wang2025visualprm,ouyang2022training,cobbe2021training,lightman2023let,hong2024orpo}.
Generative verifiers and LLM-as-judge train models to both solve and judge \cite{zhang2024generative,llm-as-judge}.
In contrast, we use reinforcement learning to train a single policy for answer and calibrated confidence with separate advantages and mutual masking, and we do not finely control the positive/negative ratio in training data; instead, we employ preference reward rather than binary reward, enabling dependable best-of-$N$ across multimodal tasks.

\section{Method}
\label{sec:method}

We propose \textbf{Advantage Decoupled Preference Optimization}, a unified reinforcement learning framework that extends GRPO to jointly learn answer generation and self verification within a single policy.
For each multimodal query, our model first generates an answer and then predicts a verification score.
At inference, we use batch decoding to sample multiple candidate answers and select the one with the highest verification score as the final output.
This unified generation and verification paradigm achieves reliable self verification and reduces inference latency.
To realize this paradigm, we develop preference verification reward to strengthen verification (\cref{sec:preference_reward}) and advantage decoupled optimization to enable joint optimization (\cref{sec:dual-adv}).

\subsection{Preliminary}
\label{sec:preliminary}

\noindent\textbf{Group Relative Policy Optimization (GRPO).}
GRPO~\cite{shao2024deepseekmath} is a reinforcement learning algorithm that optimizes language models through group-based advantage estimation.
Given a question $q$, the behavior policy $\pi_{\theta_{\text{old}}}$ samples a group of $G$ candidate responses $\{o_i\}_{i=1}^G$.
Each response $o_i{=}(o_{i,1},\dots,o_{i,|o_i|})$ is a token sequence of length $|o_i|$ that receives a sequence-level reward $R_i$.
GRPO estimates advantages by normalizing rewards within each group.
The current policy $\pi_{\theta}$ is optimized with a PPO-style~\cite{schulman2017proximal} clipped objective:
{
\small
\begin{equation}
    \begin{aligned}
        \mathcal{J}_{\text{GRPO}}(\theta)\; & = \mathbb{E}\Bigg[\frac{1}{G}\sum_{i=1}^{G}\frac{1}{|o_i|}\sum_{t=1}^{|o_i|}\Big(\min\big(r_{i,t}(\theta)\hat{A}_{i,t}, \\
        \mathrm{clip}(r_{i,t}(\theta),      & 1-\varepsilon,1+\varepsilon)\hat A_{i,t}\big)
        - \beta\, D_{\mathrm{KL}}(\pi_\theta\Vert\pi_{\mathrm{ref}})\Big)\Bigg].
    \end{aligned}
\end{equation}
}
where $r_{i,t}(\theta){=}\frac{\pi_{\theta}(o_{i,t}\mid q,o_{i,<t})}{\pi_{\theta_{\text{old}}}(o_{i,t}\mid q,o_{i,<t})}$ is the likelihood ratio,  $\varepsilon$ is the clipping parameter, $\beta$ is the KL coefficient, $D_{\text{KL}}$ is the KL regularization, $\pi_{\text{ref}}$ is the reference policy, and the group-normalized advantage is defined as $\hat{A}_{i,t}{=}\frac{R_i-\text{mean}(\{R_i\}_{i=1}^{G})}{\text{std}(\{R_i\}_{i=1}^{G})}$.

\begin{table}[t]
    \centering
    \caption{
        \textbf{Prompt for ADPO training.}
    }
    \begin{tcolorbox}
        \texttt{<image>}\texttt{<image\_pad>}\texttt{</image>} \\
        \{Question\}
        Output the thinking process in \texttt{<think></think>} and final answer in \texttt{<answer></answer>} tags.
        \\
        \textcolor{darkred}{After outputting the answer, you will act as a correctness evaluation assistant and assign a score between 0 and 1 to indicate how accurate the answer is.
            If you believe the answer is correct, the score should be close to 1; otherwise, it should be close to 0.
        } \\
        For example: \\ \texttt{<think>}reasoning process here\texttt{</think>} \\ \texttt{<answer>}answer here\texttt{</answer>} \\ \textcolor{darkred}{\texttt{<score>}score number here\texttt{</score>}}. 
        
    \end{tcolorbox}
    \label{tab:prompt}
\end{table}

\noindent\textbf{Answer Reward.}
We define the answer reward $R^a$ to evaluate generation quality by comparing model outputs against ground-truth answers.
The formulation of $R^a$ varies based on task characteristics.

For \emph{discrete tasks} with exact ground-truth answers (e.g., mathematical reasoning, agent navigation), we use correctness-based rewards:
\begin{equation}
    R^a_{\text{discrete}} = \text{match}(y, y^\ast)\in\{0,1\},
\end{equation}
where $\text{match}(\cdot,\cdot)$ performs equivalence checking (e.g., numerical equivalence for math, action sequence matching for agents), and $y$ and $y^\ast$ represent the predicted and ground truth answers, respectively.

For \emph{continuous tasks} where answer quality cannot be captured as simply correct or incorrect (e.g., visual grounding), we employ continuous rewards:
\begin{equation}
    R^a_{\text{continuous}} = \text{sim}(y, y^\ast) \in [0, 1],
\end{equation}
where $\text{sim}(\cdot, \cdot)$ denotes a similarity metric.
For visual grounding, we use Intersection-over-Union (IoU) to quantify the overlap between predicted and ground-truth regions.

\subsection{Preference Verification Reward}
\label{sec:preference_reward}

We enable self-verification by instructing the model to evaluate its response.
After generating an answer, the model outputs a verification score $s\!
    \in\![0,1]$, which represents the predicted correctness.
The prompt is provided in \cref{tab:prompt}.

We first consider a simple binary verification reward as a baseline.
The goal of self-verification is to align verification score with answer correctness so that correct answers receive higher scores and incorrect answers receive lower scores.
We define a typical \textbf{binary verification reward} $R^b$ by thresholding verification score $s$ and answer reward $R^a$:
\begin{equation}
    \label{eq:binary_reward}
    R^b = \mathbbm{1}\{{(s-\tau_s)  (R^a-\tau_a) > 0}\},
\end{equation}
The reward equals one when the predicted correctness and the actual correctness agree, while equals zero otherwise.
This objective calibrates the verifier to produce scores aligned with answer correctness.

\noindent\textbf{Class Imbalance Challenge.}
\label{par:imbalance-challenge}
The binary verification reward is susceptible to the class imbalance challenge.
As the model improves during training and generates more correct solutions, the proportion of correct samples grows substantially larger than incorrect ones.
This severe imbalance between positive and negative samples causes the verification scores to collapse to the same value, thereby eliminating the model's discriminative capability.

\begin{figure}[t]
    \centering
    \includegraphics[width=\linewidth]{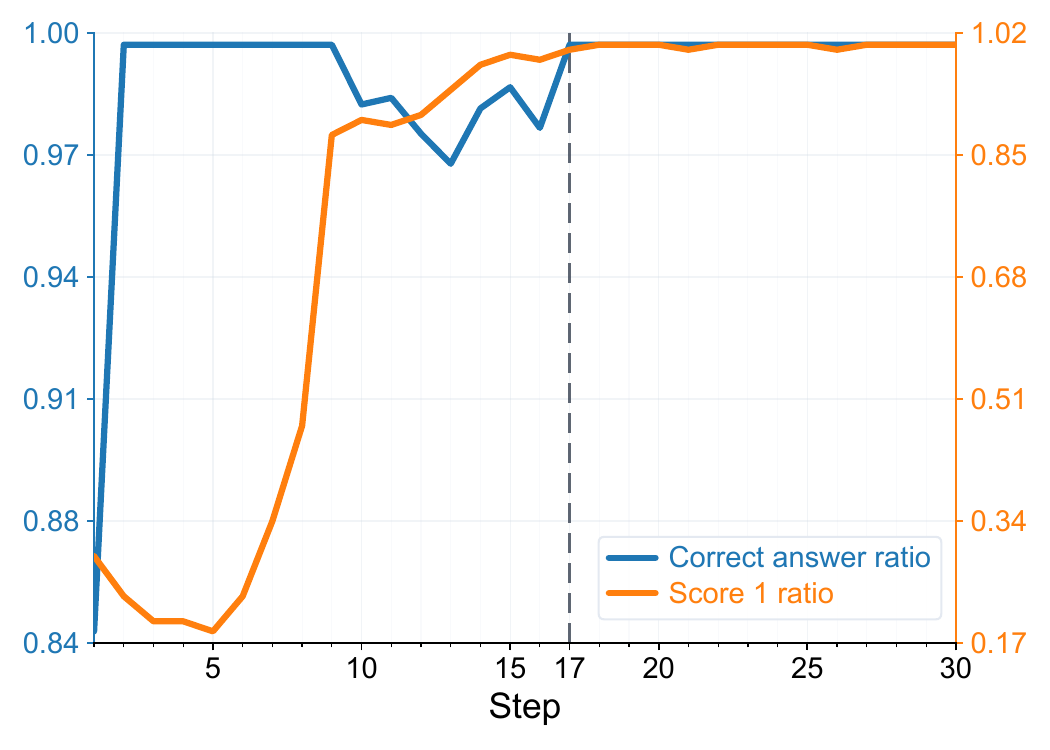}
    \caption{
        \textbf{Effect of class imbalance on the \emph{Binary Verification Reward}.}
        The \textcolor[rgb]{0.122,0.467,0.706}{\textbf{Blue}} line shows the proportion of correct answers among responses with binary verification reward = 1.
        The \textcolor[rgb]{1.000,0.498,0.055}{\textbf{Orange}} line shows the proportion of answers with verification score = 1.
    }
    \label{fig:pref_a}
\end{figure}

As shown in~\cref{fig:pref_a}, among responses that receive a binary verification reward of 1, more than 80\% answers are correct, which encourages the model to assign a verification score of 1 with increasingly high probability.
Within 17 optimization steps, the verification scores for nearly all answers converge to 1, regardless of their correctness.
When all answers receive the same verification score, their rewards collapse to the same value, the advantage becomes zero and the learning gradient vanishes.
This induces an inescapable local optimum that traps the model in producing uninformative scores and weakens its ability to distinguish correct from incorrect solutions.

\noindent\textbf{Preference Verification Reward.}
To address this limitation, we propose a preference verification reward that preserves discriminative signals even under severe class imbalance.
Instead of using fixed thresholds for verification scores, we introduce \emph{adaptive thresholds} to provide relative ranking supervision within each group.
We partition samples into contrastive sets according to answer quality and encourage the policy to assign higher verification scores to better answers and lower scores to worse ones.
Formally, for sample $i$ with verification score $s_i$ and answer reward $R^a_i$, we define the preference verification reward $R^p_i$ as:
\begin{equation}
    R_i^p = \frac{1}{\max\bigl(\lvert \mathcal{C}_i\rvert, 1\bigr)} \sum_{j \in \mathcal{C}_i} \mathbbm{1}\!
    \bigl\{(s_i - s_j)(R^a_ i - R^a_j) > 0\bigr\},
\end{equation}
where $\mathcal{C}_i$ denotes the \emph{contrastive set} for sample $i$ and contains samples with different answer qualities.
The verification reward $R_i^p$ measures \emph{ranking accuracy}, i.e., the proportion of contrastive pairs with matching verification score and answer quality orderings.
For example, when $R^a_i > R^a_j$ (sample $i$ is better), we expect $s_i > s_j$ (higher verification score), and vice versa.
The indicator function $\mathbbm{1}\!
    (s_i - s_j)(R^a_i - R^a_j) > 0$ equals 1 exactly when the ranking of verification scores and answer qualities are consistent.

For \emph{discrete tasks}, we partition each group into positive samples (correct answers) and negative samples (incorrect answers) based on $R^a$ for each sample, then define the contrastive set $\mathcal{C}_i$ for sample $i$ as:
\begin{equation}
    \mathcal{C}_i = \{ j\!\in\!\{1,\!\dots,\!
    G\} \mid R^a_j\!
    \neq\!
    R^a_i \}.
\end{equation}
Each sample is only compared against other samples in the same contrastive set, encouraging the verifier to distinguish correct from incorrect answers.
This formulation rewards the model when: (1) correct samples have verification scores above the incorrect samples' average, or (2) incorrect samples have verification scores below the correct samples' average.
Rather than treating verification as a coarse binary prediction, our method explicitly reinforces relative quality rankings between samples, thereby substantially enhancing verification capability.

\begin{figure*}[t]
    \centering
    \includegraphics[width=\linewidth]{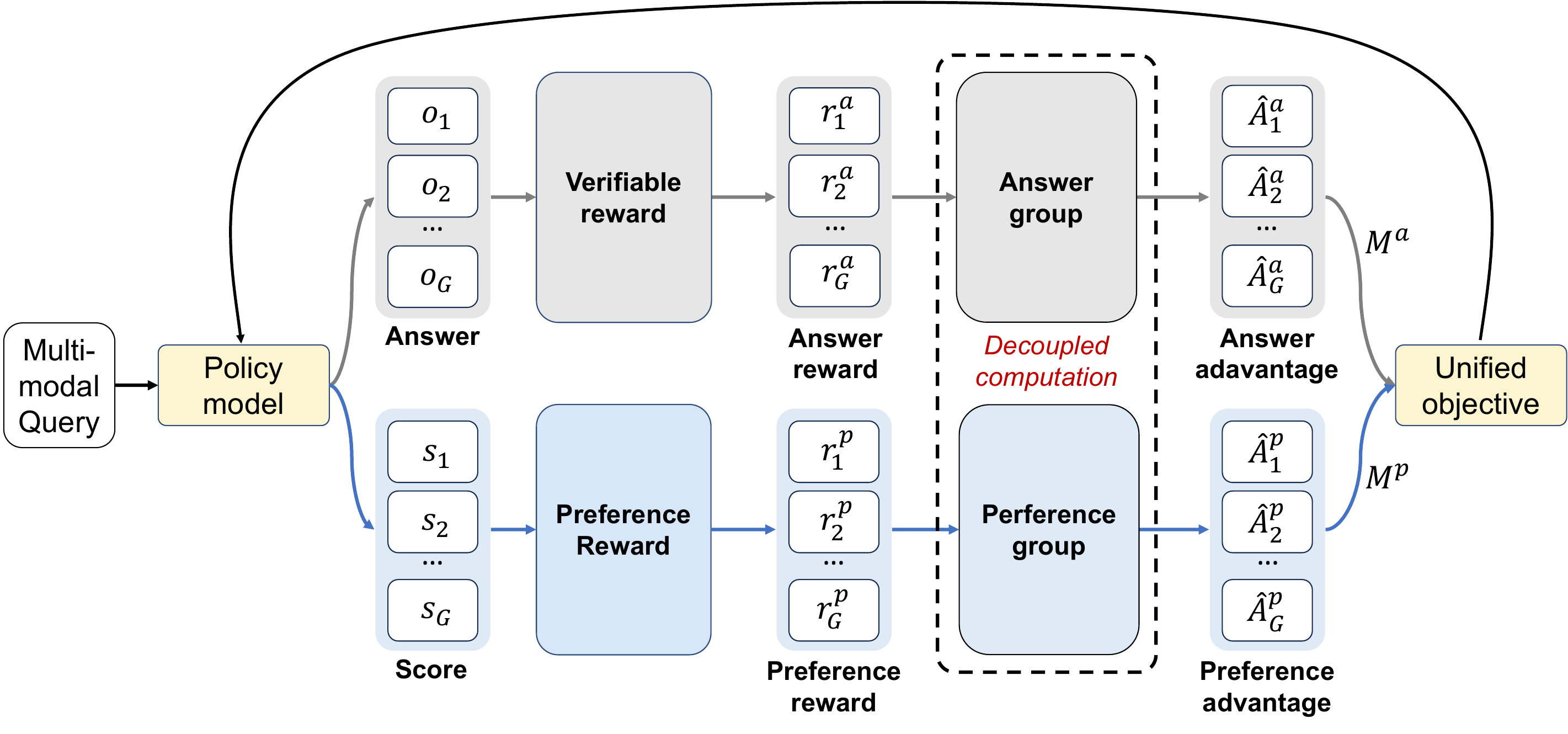}
    \caption{
        \textbf{The framework of ADPO.}
        Given a multimodal input, our unified policy produces an answer and a self-verification score to rank answer candidates.
        We design \textbf{a preference verification reward} to improve verification capability and \textbf{a decoupled optimization mechanism} to enable synergistic optimization of generation and verification.
        Preference verification reward aligns verification scores with answer correctness by providing relative ranking supervision.
        Advantage decoupled optimization computes separate advantages for generation and verification, and applies token masks to isolate gradients, thereby preventing reward hacking and reducing gradient interference between the two objectives.
    }
    \label{fig:method}
\end{figure*}

For \emph{continuous tasks}, we regard answers with similar quality as positives and others as negatives.
We impose a margin $\gamma>0$ on quality differences and define the contrastive set for sample $i$ as:
\begin{equation}
    \mathcal{C}_i = \{ j\!\in\!\{1,\!\dots,\!
    G\} \mid |R^a_j - R^a_i |\!
    >\!\gamma \}.
\end{equation}

When all rollouts are all same correct or incorrect, all assigned verification rewards are zero, resulting in no gradient updates.
This fundamentally addresses the issue where, as training progresses, an increasing proportion of samples receive identical rewards, which would otherwise lead to model outputs converging to binary values of 0 or 1.
This preference reward provides dense, contrastive supervision that maximizes \textit{quality-dependent score margins} while accommodating both discrete correctness and continuous paradigms.

Preference verification reward optimizes ranking consistency between the verification score and answer quality rather than the absolute probability scale; because the method outputs relative scores rather than calibrated probabilities, we report ranking metrics that do not depend on probability calibration, namely AUC and AP.
In ablation, preference verification reward improves both AUC and AP over the binary reward, indicating stronger alignment between scores and answer quality (see Fig.~\ref{fig:ablation_reward}).

\noindent\textbf{Discussion.}
Preference verification reward targets ranking consistency between the verification score and answer quality rather than the absolute scale of predicted probabilities.
We therefore use AUC (ROC-AUC) and AP (average precision) to show that better answers receive higher scores.
In ablation study, preference verification reward improves AUC and AP up to \textbf{0.19} over the binary reward, indicating stronger alignment between verification scores and answer quality (Fig.~\ref{fig:ablation_reward}).

\subsection{Advantage Decoupled Optimization}
\label{sec:dual-adv}
\textbf{Entangled advantage.} A straightforward approach to jointly optimize generation and verification is to simply aggregate the answer and verification rewards:
\begin{equation}
    R_\text{total}=R^a(y,y^\ast) + R^p(y, y^\ast, s).
    \label{eq:entangled_advantage}
\end{equation}
where $R_\text{total}$ denotes the aggregated reward used to compute advantages in the GRPO objective.

However, we found that naively aggregating the answer reward and the verification reward degrades generation capability.
The model learns to exploit the objective by producing an obviously incorrect answer while assigning a very low self-verification score to exploit the reward, thereby still achieving a high total reward.

This issue stems from the fact that generation and verification constitute two distinct tasks with different optimization objectives.
Specifically, answer rewards favor samples with higher response quality, while verification rewards favor samples with better calibrated verification scores, making their rewards fundamentally incompatible for simple aggregation.

\textbf{Decoupled advantage.}
To address this conflict, we decouple the advantage group by reward type and isolate gradients using disjoint token masks.
As shown in~\cref{fig:method}, we compute separate advantages within each reward group: $\hat{A}^{(a)}$ from answer group and $\hat{A}^{(p)}$ from preference group.
We then apply task-specific token masks to prevent gradient interference: $M^{a}$ covers answer generation tokens (including reasoning when present), while $M^{p}$ covers only verification score tokens.
This design prevents gradient interference.

The unified training objective is:
\begin{equation}
    \label{eq:unified_objective}
    \mathcal{J}(\theta) =
    M^{a} \odot \mathcal{J}_{\theta}(\hat{A}^{(a)}) + M^{p} \odot \mathcal{J}_{\theta}(\hat{A}^{(p)}).
\end{equation}
where $\odot$ denotes element-wise multiplication over tokens and $M^{a},M^{p}\in\{0,1\}^{T}$ are token-level masks that gate contributions and gradients to their respective regions.
By construction, improvements in answer quality are driven only by $\hat{A}^{(a)}$ on generation tokens, while calibration is shaped only by $\hat{A}^{(p)}$ on verification tokens, thereby resolving the gradient conflict in joint optimization.

\section{Experiments}
We evaluate ADPO on three multimodal domains: Math Reasoning, Visual Grounding, and Mobile Agent tasks.
Our experiments demonstrate that ADPO consistently outperforms strong baselines on standard benchmarks, achieving improvements in both task accuracy and self-verification reliability.

\subsection{Experimental Setup}
\noindent\textbf{Datasets.}
We evaluate across three domains using standard benchmarks to assess model capabilities.
(1) \textbf{Multimodal math reasoning}: we train on curated multimodal math reasoning dataset~\cite{lmmslab_multimodal_open_r1_8k_verified} and evaluate in-domain on MathVista~\cite{mathvista} and out-of-distribution on MMMU~\cite{yue2024mmmu}, reporting accuracy. (2) \textbf{Visual grounding}: we train on RefCOCO~\cite{yu2016modeling} and evaluate on ReasonSeg~\cite{lai2024lisa}, measuring cIoU for referring expression comprehension. (3) \textbf{Mobile agents}: we train separately on AndroidControl~\cite{li2024effects} and GUI Odyssey~\cite{lu2024gui} and evaluate on their official test sets for UI navigation, reporting step success rate (SR).

\noindent\textbf{Baselines.}
We compare against three primary baselines: (i) GRPO, (ii) GRPO with majority voting, and (iii) ADPO with majority voting.
For the verification baseline, we evaluate three judge variants—the base model, a GRPO-trained model as judge, and our ADPO-trained model as judge.
For multimodal mathematical tasks, we additionally compare against a baseline where the base model serves as the generator and a specialized reward model fine-tuned on mathematical data acts as the verifier.

\noindent\textbf{Implementation details.}
Unless otherwise noted, all experiments use a shared configuration: learning rate $1\times10^{-6}$, batch size $128$, group size $G{=}8$, GRPO clipping parameter $\varepsilon{=}0.2$, and KL coefficient $\beta{=}0.01$.
For \textbf{Multimodal Math Reasoning}, we fine-tune Qwen2-VL-7B~\cite{wang2024qwen2vl} for $1{,}200$ steps.
For \textbf{Visual Grounding} and \textbf{Mobile Agent}, we initialize from Qwen2.5-VL-7B~\cite{bai2025qwen2} and train for $1{,}200$ and $8{,}000$ steps, respectively.
During policy rollouts, we decode with temperature $T{=}1.0$ and $\text{top-}p{=}0.99$; at evaluation, we use $T{=}0.2$ with the same $\text{top-}p$.

\subsection{Main Results}
Our evaluation compares three generators: the base model, the GRPO-finetuned model, and the ADPO-finetuned model, each paired with majority voting as verification strategies (\cref{tab:math,tab:grounding,tab:agent}).
We conducted a comprehensive comparison of base, GRPO, and ADPO as both generators and verifiers across three domains (\cref{tab:verifier_overview}).
We report the performance under pass@1, majority voting and best-of-N evaluation protocols ($N \in {4, 8, 12}$), and find that employing ADPO as a unified generator and verifier achieves the best performance across all domains under a fixed sampling budget.
Notably, this unified approach maintains a pass@1 generation quality comparable to that of the GRPO model.

\begin{table}[!htbp]
    \centering
    \caption{
        \textbf{Performance on MathVista~\cite{mathvista} and MMMU~\cite{yue2024mmmu}.}
        We adopt Qwen2-VL-7B~\cite{wang2024qwen2vl} as the base model and use majority voting for both the base and GRPO models.
        We report accuracy (\%) and highlight the best results in \textbf{bold}.
    }
    \label{tab:math}
    \resizebox{\linewidth}{!}{
        \setlength{\tabcolsep}{2pt}
        \begin{tabular}{c|ccc|ccccccc}
            \toprule
            \multirow{2}{*}{\textbf{Method}}   & \multicolumn{3}{c|}{\textbf{MathVista (In-domain)}} & \multicolumn{7}{c}{\textbf{MMMU (OOD)}}                                                                                                                                 \\
                                               & GVQA                                                & MVQA                                    & \textbf{ALL}  & ARD           & BUS           & HEM           & HSS           & SCI           & TEN           & \textbf{ALL}  \\
            \midrule
            \rowcolor{backblue2} \multicolumn{11}{c}{\textit{Sample 1}}                                                                                                                                                                                                        \\
            \midrule
            R1-VL-7B~\cite{zhang2025r1}        & -                                                & -                                    & 63.5          & -             & -             & -             & -             & -             & -             & -             \\
            \noalign{\vskip 2pt} \cdashline{1-11} \noalign{\vskip 2pt}
            Base & 68.9                                                & 48.5                                    & 57.9          & \textbf{67.5} & 39.1          & 49.3          & 69.0          & 33.9          & 36.7          & 47.1          \\
            GRPO                               & \textbf{69.8}                                       & 55.7                                    & 62.2          & 65.0          & 45.9          & 48.2          & 68.2          & \textbf{35.9} & \textbf{39.8} & \textbf{48.7} \\
            ADPO                               & 68.7                                                & \textbf{57.0}                           & \textbf{62.4} & 63.1          & \textbf{46.2} & \textbf{50.2} & \textbf{71.1} & 33.3          & 35.3          & 47.7          \\
            \midrule
            \rowcolor{backblue2} \multicolumn{11}{c}{\textit{Sample 4}}                                                                                                                                                                                                        \\
            \midrule
            MM-Verifier~\cite{sun2025mm}       & 67.0                                                & 53.7                                    & 59.8          & -             & -             & -             & -             & -             & -             & -             \\
            \noalign{\vskip 2pt} \cdashline{1-11} \noalign{\vskip 2pt}
            Base                       & 65.7                                                & 51.9                                    & 58.2          & 66.7          & 47.3          & 50.7          & 65.8          & 34.0          & 38.1          & 48.6          \\
            GRPO                       & 69.8                                                & 58.0                                    & 63.4          & 65.8          & 44.7          & 50.0          & \textbf{70.0}          & \textbf{42.0}          & 36.7          & 49.4          \\
            ADPO                        & \textbf{71.3}                                                & \textbf{59.3}                                    & \textbf{64.8}          & \textbf{68.3}          & \textbf{48.0}          & \textbf{52.0}          & 69.2          & 39.3          & \textbf{39.5}          & \textbf{50.8}          \\
            \midrule
            \rowcolor{backblue2} \multicolumn{11}{c}{\textit{Sample 8}}                                                                                                                                                                                                        \\
            \midrule
            MM-Verifier~\cite{sun2025mm}       & 68.5                                                & 57.4                                    & 62.5          & -             & -             & -             & -             & -             & -             & -             \\
            \noalign{\vskip 2pt} \cdashline{1-11} \noalign{\vskip 2pt}
            Base                       & 68.0                                                & 53.3                                    & 60.1          & \textbf{68.3}          & 50.0          & 53.3          & 68.3          & 32.7          & 36.7          & 49.4          \\
            GRPO                       & 70.4                                                & 56.5                                    & 62.9          & 66.7          & 48.7          & 51.3          & \textbf{74.2}          & \textbf{42.7}          & 36.7          & 51.1          \\
            ADPO                         & \textbf{72.2}                                                & \textbf{58.9}                                    & \textbf{65.0}          & 65.8          & \textbf{54.0}          & \textbf{54.7}          & 66.7          & 40.7          & \textbf{41.0}          & \textbf{52.1}          \\
            \midrule
            \rowcolor{backblue2} \multicolumn{11}{c}{\textit{Sample 12}}                                                                                                                                                                                                       \\
            \midrule
            MM-Verifier~\cite{sun2025mm}       & 70.4                                                & 58.7                                    & 64.1          & -             & -             & -             & -             & -             & -             & -             \\
            \noalign{\vskip 2pt} \cdashline{1-11} \noalign{\vskip 2pt}
            Base                       & 67.4                                                & 55.0                                    & 60.7          & \textbf{69.2}          & 52.0          & 50.7          & 70.8          & 38.0          & 36.7          & 50.7          \\
            GRPO                       & 70.7                                                & 57.2                                    & 63.4          & 64.2          & 50.0          & 51.3          & 73.3          & \textbf{43.3}          & 39.5          & 51.7          \\
            ADPO                        & \textbf{71.7}                                                & \textbf{59.8}                                    & \textbf{65.3}          & 67.5          & \textbf{53.3}          & \textbf{54.0}          & \textbf{71.7}          & 38.7          & \textbf{40.5}          & \textbf{52.3}          \\
            \bottomrule
        \end{tabular}
    }
\end{table}

\begin{table}[!htbp]
    \centering
    \caption{
        \textbf{Performance on ReasonSeg~\cite{lai2024lisa}.}
        We use Qwen2.5-VL-7B~\cite{bai2025qwen2} as the base model. 
    }
    \label{tab:grounding}
    \resizebox{\linewidth}{!}{
        \begin{tabular}{c|ccc|ccc|ccc}
            \toprule
            \multirow{2}{*}{\textbf{Method}}  & \multicolumn{3}{c|}{\textbf{Short query}} & \multicolumn{3}{c|}{\textbf{Long query}} & \multicolumn{3}{c}{\textbf{Overall}}                                                                                                 \\
                                              & \textbf{gIoU}                             & \textbf{cIoU}                            & \textbf{ACC}                         & \textbf{gIoU} & \textbf{cIoU} & \textbf{ACC}  & \textbf{gIoU} & \textbf{cIoU} & \textbf{ACC}  \\
            \midrule
            \rowcolor{backblue2} \multicolumn{10}{c}{\textit{Sample 1}}                                                                                                                                                                                                     \\
            \midrule
            LISA-7B~\cite{lai2024lisa}        & 47.1                                      & 48.5                                     & -                                    & 49.2          & 48.9          & -             & 48.7          & 48.8          & -             \\
            SegLLM~\cite{wang2024segllm}      & -                                         & -                                        & -                                    & -             & 54.2          & -             & -             & 48.4          & -             \\
            Seg-Zero-7B~\cite{liu2025seg}     & -                                         & -                                        & -                                    & -             & -             & -             & 57.5          & 52.0          & -             \\
            VLM-R1~\cite{shen2025vlm}         & -                                         & -                                        & -                                    & -             & -             & -             & -             & -             & 63.1          \\
            \noalign{\vskip 2pt} \cdashline{1-10} \noalign{\vskip 2pt}
            Base & 49.5                                      & 53.0                                     & 67.0                                 & 56.8          & 57.5          & 68.5          & 56.3          & 57.2          & 68.4          \\
            GRPO                              & 51.8                                      & \textbf{55.5}                                     & 67.9                                 & 59.1          & \textbf{59.7}          & 71.3          & \textbf{58.6}          & \textbf{59.5}          & 71.1          \\
            ADPO                              & \textbf{51.7}                                      & 54.8                                     & \textbf{68.0}                        & \textbf{60.2}          & 59.4          & \textbf{71.9} & 58.1          & 59.1          & \textbf{71.7} \\
            \midrule
            \rowcolor{backblue2} \multicolumn{10}{c}{\textit{Sample 4}}                                                                                                                                                                                                     \\
            \midrule
            Base                      & 47.8                                      & 52.0                                     & 66.0                                 & 57.3          & 57.9          & 69.3 & 56.7          & 57.5          & 69.1 \\
            GRPO                      & \textbf{54.5}                                      & \textbf{57.0}                                     & \textbf{68.0}                                 & 58.8          & 59.5          & 72.1          & 58.5          & 59.4          & 71.8          \\
            ADPO                        & 52.2                                      & 55.1                                     & 67.0                                 & \textbf{61.0}          & \textbf{61.5}          & \textbf{73.3} & \textbf{60.5}          & \textbf{61.1}          & \textbf{72.9} \\
            \midrule
            \rowcolor{backblue2} \multicolumn{10}{c}{\textit{Sample 8}}                                                                                                                                                                                                     \\
            \midrule
            Base                      & 47.8                                      & 51.4                                     & 63.1                        & 57.2          & 57.8          & 69.2          & 56.6          & 57.4          & 68.8          \\
            GRPO                      & 52.0                                      & 55.6                                     & \textbf{68.0}                        & 59.2          & 59.9          & 72.0          & 58.7          & 59.6          & 71.7          \\
            ADPO                       & \textbf{53.2}                                      & \textbf{56.0}                                     & 67.0                                 & \textbf{60.9}          & \textbf{61.5}          & \textbf{73.7} & \textbf{60.4}          & \textbf{61.2}          & \textbf{73.5} \\
            \midrule
            \rowcolor{backblue2} \multicolumn{10}{c}{\textit{Sample 12}}                                                                                                                                                                                                    \\
            \midrule
            Base                      & 50.2                                      & 53.7                                     & 66.0                                 & 57.2          & 57.8          & 69.3 & 56.8          & 57.6          & 69.1 \\
            GRPO                      & \textbf{55.6}                                      & \textbf{58.1}                                     & \textbf{69.9}                        & 58.8          & 59.5          & 72.2          & 58.6          & 59.4          & 72.0          \\
            ADPO                       & 53.9                                      & 56.2                                     & 67.0                                 & \textbf{61.3}          & \textbf{62.0}          & \textbf{73.6} & \textbf{60.9}          & \textbf{61.6}          & \textbf{73.2} \\
            \bottomrule
        \end{tabular}
    }
\end{table}

\begin{table}[!htbp]
    \centering
    \caption{
        \textbf{Performance on AndroidControl~\cite{li2024effects} and GUI Odyssey~\cite{lu2024gui}.}
        We adopt Qwen2.5-VL-7B as base model and report type accuracy, grounding accuracy and step success rate (SR).
    }
    \label{tab:agent}
    \resizebox{\linewidth}{!}{
        \begin{tabular}{c|ccc|ccc}
            \toprule
            \multirow{2}{*}{\textbf{Generator}}        & \multicolumn{3}{c|}{\textbf{AndroidControl}} & \multicolumn{3}{c}{\textbf{GUI Odyssey}}                                                                 \\
                                                       & Type                                         & Grounding                                & \textbf{SR}   & Type          & Grounding     & \textbf{SR}   \\
            \midrule
            \rowcolor{backblue2} \multicolumn{7}{c}{\textit{Sample 1}}                                                                                                                                           \\
            \midrule
            UI-TARS-7B~\cite{qin2025uitars}            & 83.7                                         & -                                        & 72.5          & 86.1          & -             & 67.9          \\
            SpiritSight-8B~\cite{huang2025spiritsight} & -                                            & -                                        & 68.1          & -             & -             & 75.8          \\
            AgentCPM-GUI-8B~\cite{zhang2025agentcpm}   & 77.7                                         & -                                        & 69.2          & 90.8          & -             & 75.0          \\
            \noalign{\vskip 2pt} \cdashline{1-7} \noalign{\vskip 2pt}
            Base          & 82.2                                         & 73.6                                     & 61.3          & 81.1          & 61.4          & 52.8          \\
            GRPO                                       & \textbf{86.0}                                & \textbf{76.9}                            & \textbf{71.0} & 93.1          & \textbf{83.9} & \textbf{79.8} \\
            ADPO                                       & 85.8                                         & 76.2                                     & 70.9          & \textbf{94.2} & 82.5          & 79.7          \\
            \midrule
            \rowcolor{backblue2} \multicolumn{7}{c}{\textit{Sample 4}}                                                                                                                                           \\
            \midrule
            Base                               & 76.3                                         & 68.1                                     & 56.0          & 76.9          & 55.3          & 46.5          \\
            GRPO                               & 85.5                                         & 77.2                                     & 71.0          & 94.7 & 83.9          & 81.3          \\
            ADPO                                & \textbf{86.3}                                         & \textbf{79.5}                            & \textbf{72.7} & \textbf{94.7} & \textbf{84.5} & \textbf{81.6} \\
            \midrule
            \rowcolor{backblue2} \multicolumn{7}{c}{\textit{Sample 8}}                                                                                                                                           \\
            \midrule
            Base                               & 78.7                                         & 68.8                                     & 58.3          & 76.7          & 55.4          & 46.6          \\
            GRPO                               & 85.6                                         & 76.9                                     & 70.8          & 94.6          & 84.4          & 81.5          \\
            ADPO                                 & \textbf{86.4}                                & \textbf{78.7}                            & \textbf{72.7} & \textbf{94.8} & \textbf{84.7} & \textbf{81.7} \\
            \midrule
            \rowcolor{backblue2} \multicolumn{7}{c}{\textit{Sample 12}}                                                                                                                                          \\
            \midrule
            Base                               & 78.9                                         & 68.7                                     & 58.3          & 76.9          & 55.5          & 46.9          \\
            GRPO                               & 85.6                                         & 77.4                                     & 71.1          & \textbf{94.5} & 84.0          & 81.1          \\
            ADPO                                & \textbf{86.3}                                         & \textbf{78.9}                            & \textbf{72.9} & 94.4          & \textbf{84.5} & \textbf{81.4} \\
            \bottomrule
        \end{tabular}
    }
    \vspace{-1em}
\end{table}

\noindent\textbf{Multimodal Math Reasoning.}
On MathVista (\cref{tab:math}), ADPO's best-of-N performance steadily improves as the sample size increases, achieving 64.8\% (N=4), 65.0\% (N=8), and 65.3\% (N=12).
This approach consistently outperforms GRPO with majority voting by +1.4, +2.1, and +1.9 percentage points at the respective sample sizes, while also exceeding MM-Verifier by +5.0, +2.5, and +1.2 percentage points at these same budgets.
On MMMU, ADPO's pass@1 is competitive at 47.7\% (vs.
48.7\% for GRPO), while leading several categories.
Under best-of-N, ADPO establishes clear advantages, reaching 50.8\%, 52.1\%, and 52.3\% at N=4, 8, and 12, surpassing GRPO (majority) by +1.4, +1.0, and +0.6 points.
These results show that ADPO's unified generation–verification training strengthens sample selection and scales effectively with N, yielding higher in-domain accuracy and consistent OOD gains.

\noindent\textbf{Visual Grounding.}
On ReasonSeg (\cref{tab:grounding}), ADPO is competitive at pass@1, with overall cIoU 59.1 (vs.
59.5 for GRPO and 57.2 for the base) and the highest overall ACC of 71.7 (vs. 71.1 and 68.4).
Under best-of-N, using ADPO as a unified generator–verifier to select candidates yields overall cIoU of 61.1/61.2/61.6 at N=4/8/12, exceeding GRPO (majority voting) by +1.7/ +1.6/+2.2 and the base (majority voting) by +3.6/+3.8/+4.0; overall ACC reaches 72.9/73.5/73.2 at the same budgets.
Gains persist on long queries, where ADPO attains pass@1 gIoU/ACC of 60.2/71.9, indicating more robust localization and self-verification that continue to benefit from larger sample budgets.

\noindent\textbf{Mobile Agent.}
On AndroidControl (\cref{tab:agent}), ADPO reaches 70.9\% pass@1 SR, comparable to GRPO (71.0\%).
With best-of-N self-verification, ADPO remains stable at 72.7/72.7/72.9 for N=4/8/12, surpassing GRPO (majority) by +1.7/+1.9/+1.8 and the base model by +16.7/+14.4/+14.6 points.
On GUI Odyssey, ADPO attains 79.7\% pass@1 (vs.
79.8\% for GRPO) and, under best-of-N, improves over GRPO (majority) at N=4/8/12.

\begin{table}[t]
    \centering
    \caption{
        \textbf{Performance of different generator–verifier settings on MathVista~\cite{mathvista}, ReasonSeg~\cite{lai2024lisa} and AndroidControl~\cite{li2024effects}.}
    }
    \label{tab:verifier_overview}
    \resizebox{\linewidth}{!}{
        \setlength{\tabcolsep}{4pt}
        \begin{tabular}{c|ccc|ccc|ccc}
            \toprule
           \multirow{2}{*}{\diagbox{\footnotesize\textbf{Generator}}{\footnotesize\textbf{Verifier}}} & \multicolumn{3}{c|}{\textbf{MathVista}} & \multicolumn{3}{c|}{\textbf{ReasonSeg}} & \multicolumn{3}{c}{\textbf{AndroidControl}}                                                                                             \\
                                                                                                                                         & Base                                    & GRPO                                        & ADPO           & Base & GRPO & ADPO           & Base & GRPO & ADPO          \\
            \midrule
            \rowcolor{backblue2} \multicolumn{10}{c}{\textit{Sample 4}}                                                                                                                                                                                                                                                              \\
            \midrule
            Base                                                                                                                           & 55.7                                    & 55.5                                        & 56.4            & 57.1 & 57.7 & 57.7            & 52.5 & 57.7 & 60.7          \\
            GRPO                                                                                                                           & 62.4                                    & 62.1                                        & 62.0            & 60.2 & 59.5 & 60.9            & 71.0 & 70.8 & 71.2          \\
            ADPO                                                                                                                           & 61.5                                    & 62.1                                        & \textbf{64.8}   & 59.6 & 60.3 & \textbf{61.1}   & 71.0 & 72.0 & \textbf{72.7} \\
            \midrule
            \rowcolor{backblue2} \multicolumn{10}{c}{\textit{Sample 8}}                                                                                                                                                                                                                                                              \\
            \midrule
            Base                                                                                                                           & 57.0                                    & 56.4                                        & 56.5            & 56.9 & 57.0 & 57.9           & 54.3 & 61.0 & 64.7          \\
            GRPO                                                                                                                           & 60.7                                    & 60.8                                        & 60.5            & 60.4 & 60.4 & 61.1           & 71.0 & 70.9 & 71.4          \\
            ADPO                                                                                                                           & 62.3                                    & 62.3                                        & \textbf{65.0}   & 59.9 & 60.5 & \textbf{61.2}  & 70.8 & 71.4 & \textbf{72.7} \\
            \midrule
            \rowcolor{backblue2} \multicolumn{10}{c}{\textit{Sample 12}}                                                                                                                                                                                                                                                             \\
            \midrule
            Base                                                                                                                           & 56.9                                    & 56.3                                        & 55.0            & 57.4 & 57.6 & 57.8           & 53.6 & 60.7 & 64.5          \\
            GRPO                                                                                                                           & 62.5                                    & 62.5                                        & 61.8            & 59.7 & 59.6 & 61.3           & 71.4 & 70.9 & 71.5          \\
            ADPO                                                                                                                           & 63.0                                    & 63.5                                        & \textbf{65.3}   & 60.7 & 60.7 & \textbf{61.6}  & 71.6 & 71.9 & \textbf{72.9} \\
            \bottomrule
        \end{tabular}
    }
\end{table}

\noindent\textbf{Discussion.}
Using ADPO purely as a verifier improves selection across generators and domains (\cref{tab:verifier_overview}).
Paired with the ADPO generator, it achieves the strongest outcomes at all budgets (e.g., MathVista 64.8/65.0/65.3, ReasonSeg 61.1/61.2/61.6, AndroidControl 72.7/72.7/72.9 at N=4/8/12).
When judging outputs from base or GRPO generators, ADPO is typically best on ReasonSeg and AndroidControl and competitive on MathVista.
These findings support that decoupled advantage training calibrates verification scores well and translates into consistent best-of-N gains while preserving pass@1 performance.

Empirical results demonstrate that ADPO enables effective self-verification without degrading generation quality.
ADPO strengthens verification capability without sacrificing pass@1 performance, remaining comparable to GRPO across three domains—MathVista 62.4\% vs 62.2\% (+0.2\%), ReasonSeg 59.1\% vs 59.5\% (-0.4\%), and AndroidControl 70.9\% vs 71.0\% (-0.1\%).
The unified training framework yields models that both generate high-quality outputs and reliably select the best among multiple candidates, thereby achieving superior best-of-N performance across diverse multimodal benchmarks under rigorously consistent evaluation protocols.

\subsection{Ablation Studies}

We conduct comprehensive ablation studies to analyze the key components of our decoupled advantage preference optimization framework.

\begin{figure}[t]
    \centering
    \begin{minipage}[b]{0.5\linewidth}
        \centering
        \includegraphics[width=\linewidth]{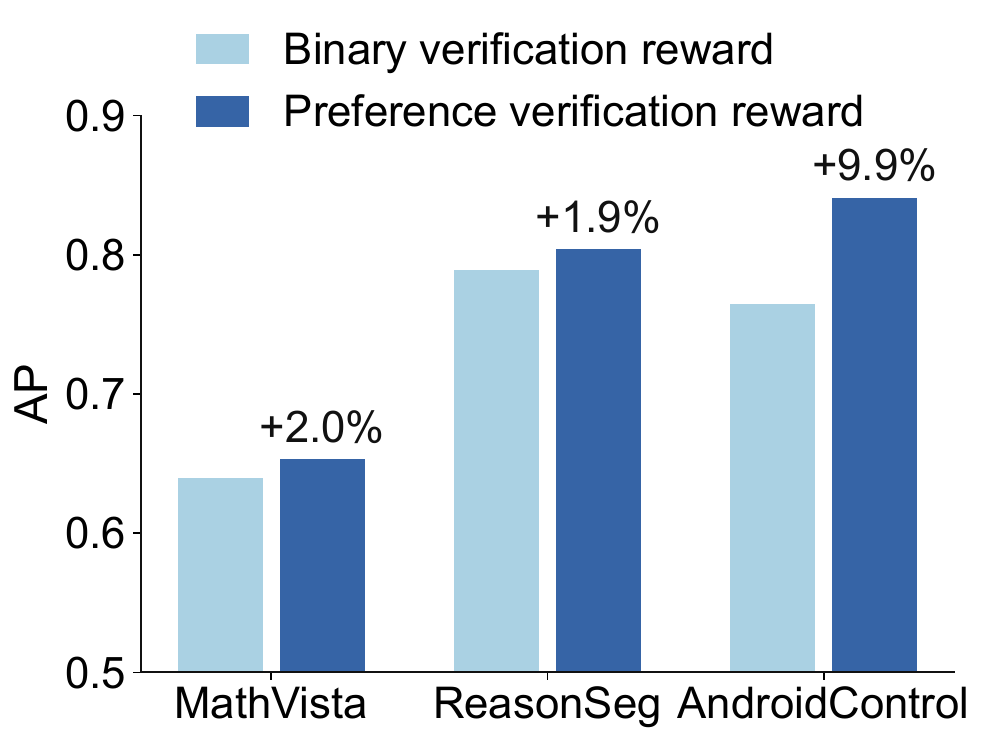}
        \subcaption{AP improvement}\label{fig:reward_ap}
    \end{minipage}\hfill
    \begin{minipage}[b]{0.5\linewidth}
        \centering
        \includegraphics[width=\linewidth]{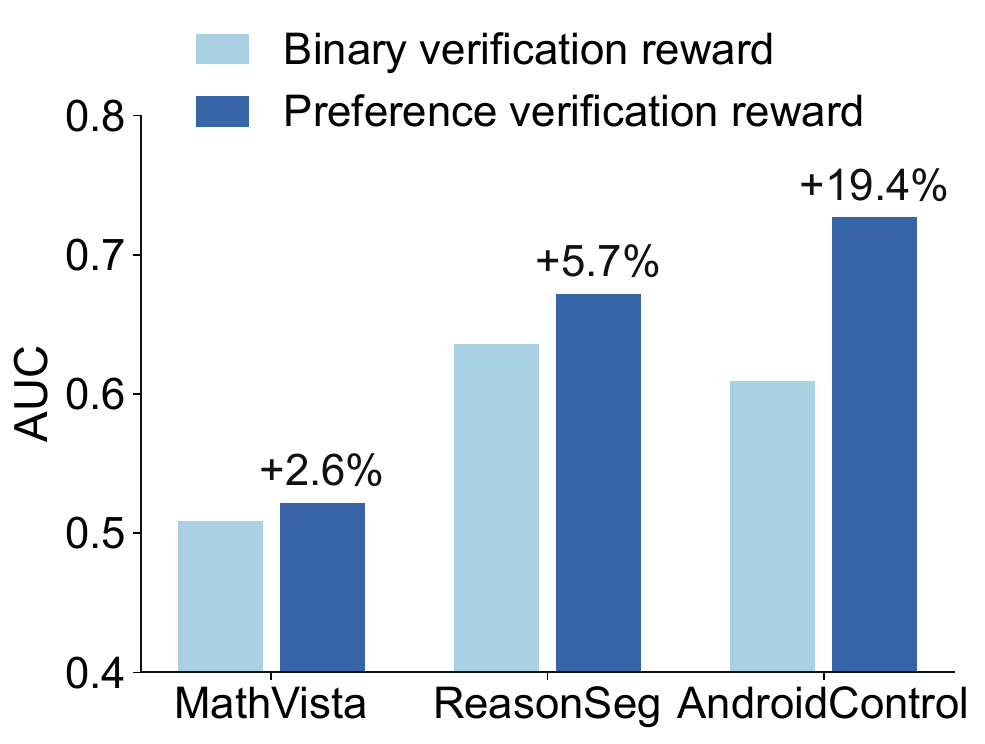}
        \subcaption{AUC improvement}\label{fig:reward_auc}
    \end{minipage}
    \caption{
        \textbf{Ablation of binary verification reward and preference verification reward.}
    }
    \label{fig:ablation_reward}
    \vspace{-1em}
\end{figure}

\begin{figure}[t]
    \centering
    \begin{minipage}[b]{0.5\linewidth}
        \centering
        \includegraphics[width=\linewidth]{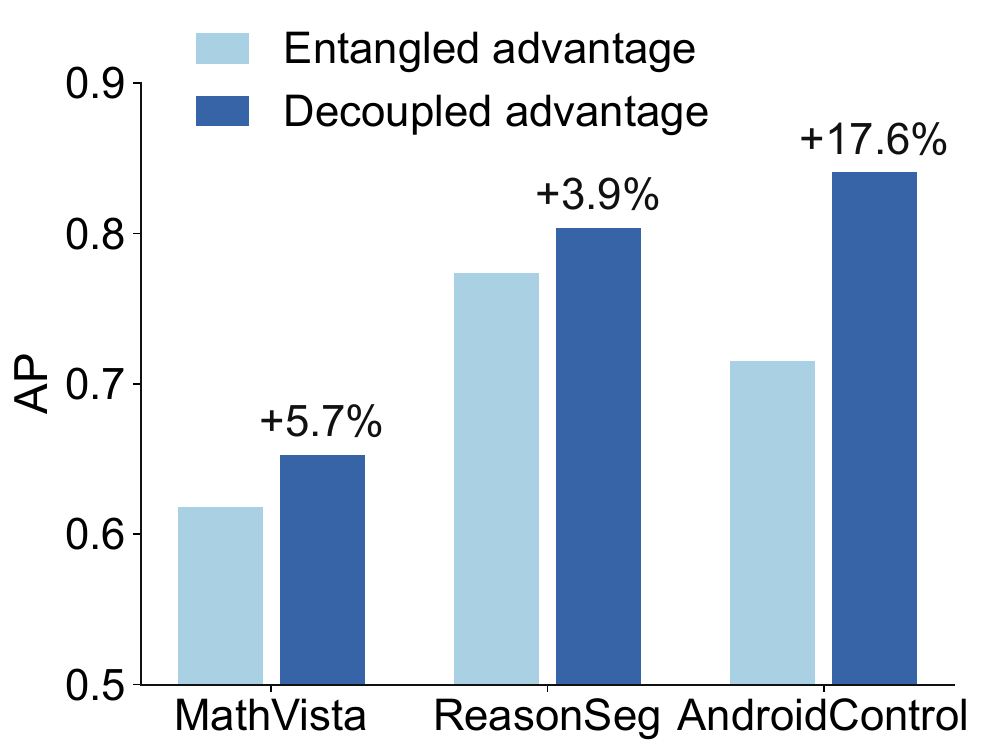}
        \subcaption{AP improvement}\label{fig:adv_ap}
    \end{minipage}\hfill
    \begin{minipage}[b]{0.5\linewidth}
        \centering
        \includegraphics[width=\linewidth]{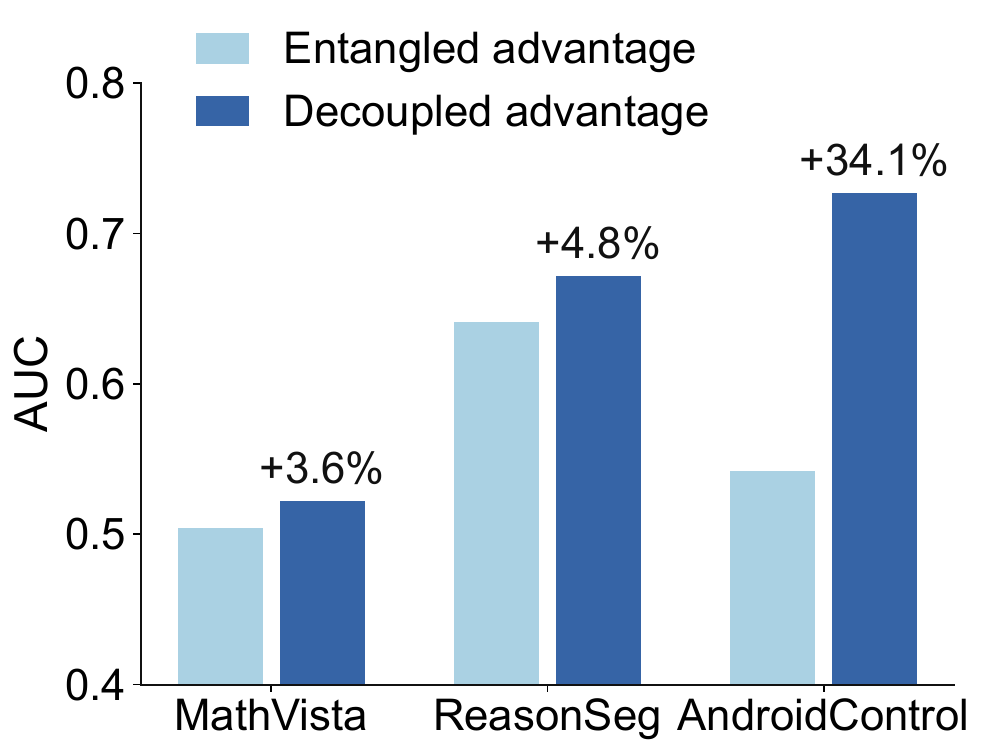}
        \subcaption{AUC improvement}\label{fig:adv_auc}
    \end{minipage}
    \caption{
        \textbf{Ablation of entangled and decoupled advantage.}
        Entangled and decoupled correspond to models trained with \emph{entangled advantage} in \cref{eq:entangled_advantage} and \emph{decoupled advantage} in \cref{eq:unified_objective}.
    }
    \label{fig:ablation_adv}
\end{figure}

\begin{figure}[t]
    \centering
    \begin{minipage}[b]{0.49\linewidth}
        \centering
        \includegraphics[width=\linewidth]{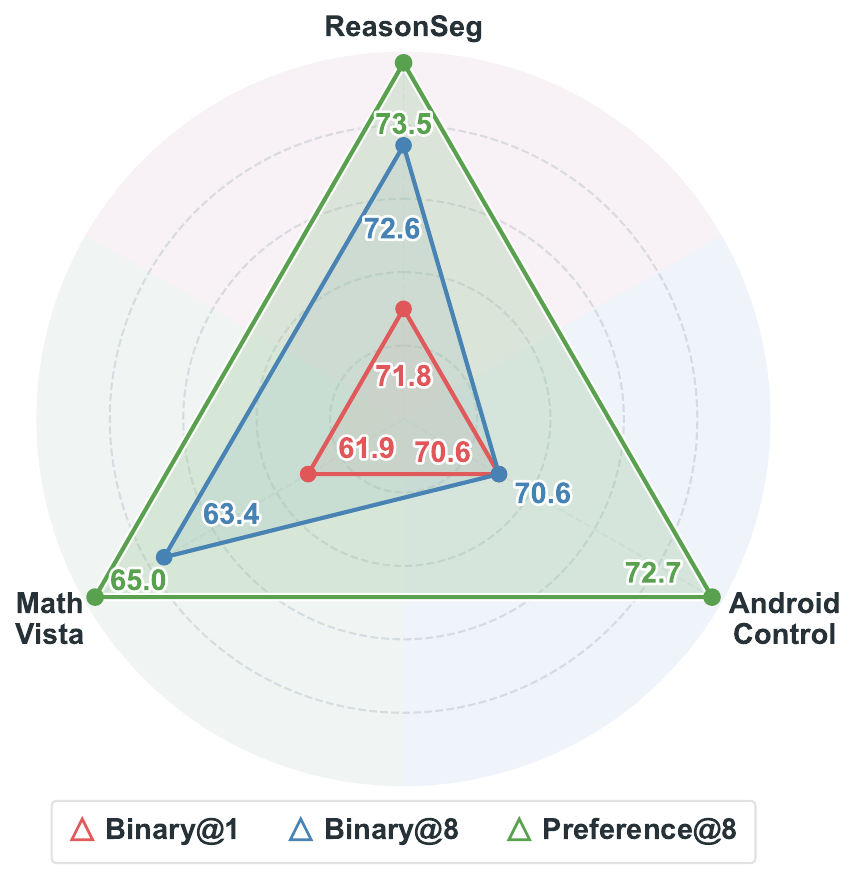}
        \subcaption{Preference verification reward}\label{fig:reward_radar}
    \end{minipage}
    \begin{minipage}[b]{0.49\linewidth}
        \centering
        \includegraphics[width=\linewidth]{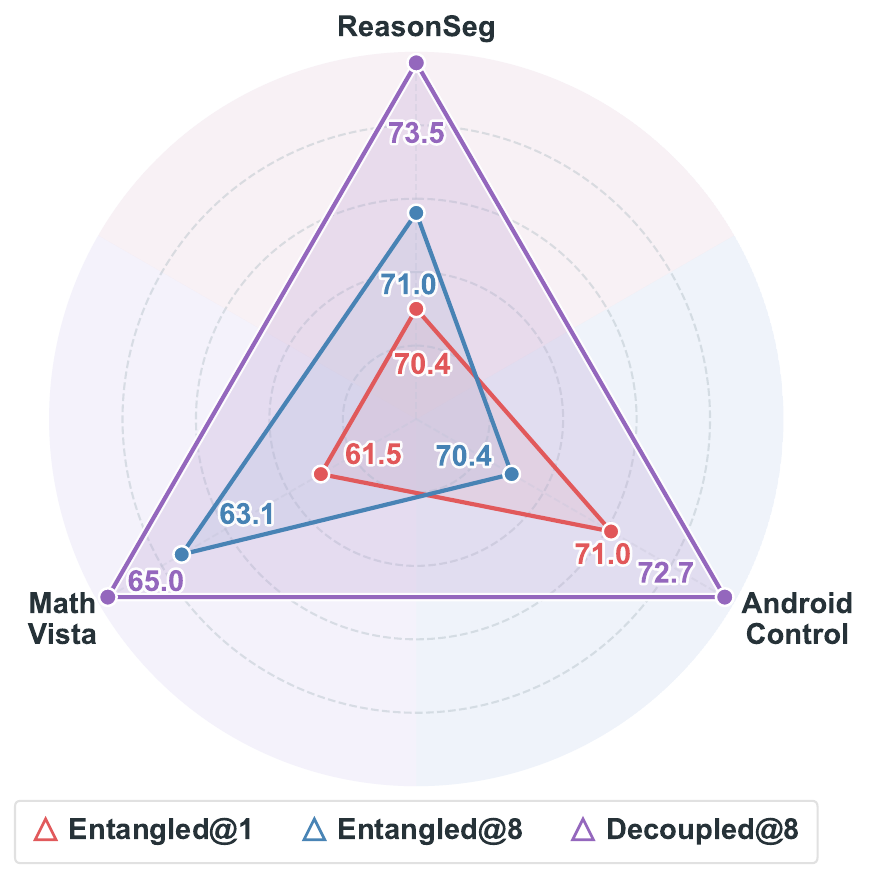}
        \subcaption{Decoupled advantage}\label{fig:adv_radar}
    \end{minipage}
    \caption{
        \textbf{Ablation of preference verification reward and advantage decoupled optimization.}
        Binary and preference denote models trained with \emph{binary verification reward} and \emph{preference verification reward}, respectively.
        The suffix @k indicates evaluation with k samples.
    }
    \label{fig:ablation_radar}
    \vspace{-1em}
\end{figure}

\noindent\textbf{Effect of preference reward.}
Figure~\ref{fig:ablation_reward} shows the impact of preference reward compared to binary reward across all three domains.
The preference formulation consistently improves both task performance and self-verification quality.
For mathematical reasoning, we observe +1.6\% improvement in best@8 performance (\cref{fig:ablation_radar}) and +1.3\% improvement in average precision (AP).
The benefits are even more pronounced for self-verification metrics, with AUC improvements of +1.3\%, +3.6\%, and +11.8\% for math, grounding, and agent tasks respectively.
This demonstrates that preference reward provide more stable training signals and better calibrated confidence scores, particularly important under the naturally imbalanced positive/negative distributions in self-verification learning.

\noindent\textbf{Effect of decoupled advantage.}
Figure~\ref{fig:ablation_adv} illustrates the contribution of our decoupled advantage computation with mutual loss masking compared to simple reward aggregation.
Decoupled advantage consistently outperform entangled advantages across all domains, with particularly significant improvements in self-verification quality.
For GUI agent tasks, decoupled advantage achieve +2.8\% improvement in best@8 performance (\cref{fig:ablation_radar}) and substantial gains in AUC +18.5 This validates our hypothesis that separating gradient flows between content generation and self-judgment prevents reward hacking and enables more effective optimization of both objectives.

\noindent\textbf{Margin parameter analysis.}
Table~\ref{tab:ablation_margin} analyzes the effect of margin parameter $\gamma$ in continuous preference reward computation for visual grounding tasks.
We find that $\gamma=0.1$ provides the optimal balance, achieving 73.5\% overall accuracy.
Too small margins ($\gamma=0.025$) may not provide sufficient discrimination between similar quality outputs, while too large margins ($\gamma \geq 0.2$) may be overly restrictive and reduce the density of preference signals.
This hyperparameter study confirms the importance of carefully tuning the preference threshold for optimal performance.

\noindent\textbf{Discussion.}
Table~\ref{tab:cost} summarizes a comparison on MathVista under a best-of-8 setting (sample size $N{=}8$).
Our unified policy (ADPO) achieves 65.0\% accuracy with a 2.6s latency per sample, improving over GRPO paired with majority voting (62.9\%, 2.1s) and GRPO-as-judge (60.8\%, 5.6s).
This shows that, at the same sampling budget, the unified generator–verifier trained by ADPO delivers stronger verification capability than GRPO while substantially reducing inference time relative to GRPO-as-judge.
In practice, this unified policy model lowers overall system complexity and deployment overhead, providing a more cost-effective path to competitive best-of-$N$ performance on MathVista.

\begin{table}[t]
    \centering
    \caption{
        \textbf{Ablation of the margin $\gamma$ for preference verification reward on ReasonSeg.}
    }
    \label{tab:ablation_margin}
    \resizebox{\linewidth}{!}{
        \begin{tabular}{c|ccc|ccc|ccc}
            \toprule
            \multirow{2}{*}{\textbf{$\gamma$}} & \multicolumn{3}{c|}{\textbf{Short query}} & \multicolumn{3}{c|}{\textbf{Long query}} & \multicolumn{3}{c}{\textbf{Overall}}                                                                                                 \\
                                               & gIoU                                      & cIoU                                     & ACC                                  & gIoU          & cIoU          & ACC           & gIoU          & cIoU          & ACC           \\
            \midrule
            0.025                              & \textbf{53.7}                             & 56.5                                     & \textbf{69.9}                        & 58.1          & 58.9          & 71.3          & 57.8          & 58.8          & 71.1          \\
            0.050                              & 52.6                                      & 54.4                                     & 63.1                                 & 60.2          & 61.0          & 73.3          & 59.8          & 60.5          & 72.7          \\
            \rowcolor{Grayheavy}
            \textbf{0.100}                     & 53.2                                      & 56.0                                     & 67.0                                 & \textbf{60.9} & \textbf{61.5} & \textbf{73.7} & \textbf{60.4} & \textbf{61.2} & \textbf{73.5} \\
            0.200                              & 53.2                                      & 55.7                                     & 66.0                                 & 59.9          & 60.7          & 72.7          & 59.6          & 60.4          & 72.3          \\
            0.250                              & \textbf{53.7}                             & \textbf{56.8}                            & 68.9                                 & 59.7          & 60.4          & 72.5          & 59.3          & 60.2          & 72.3          \\
            \bottomrule
        \end{tabular}
    }
\end{table}

\begin{table}[t]
    \footnotesize
    \centering
    \setlength{\tabcolsep}{4pt}
    \caption{
        \textbf{Comparison of unified and separate verification.}
        \emph{GRPO:}
        GRPO post-trained model as generator.
        \emph{+Major:} majority voting as verifier.
        \emph{+Judge:}
        GRPO post-trained model as verifier.
    }
    \label{tab:unified-vs-two}
    \begin{tabular}{ccc}
        \toprule
        Method     & MathVista Acc. $\uparrow$ & Latency (s) $\downarrow$ \\
        \midrule
        GRPO+Major & 62.9                      & \textbf{2.1}             \\
        GRPO+Judge & 60.8                      & 5.6                      \\
        ADPO       & \textbf{65.0}             & 2.6                      \\
        \bottomrule
    \end{tabular}
    \label{tab:cost}
    \vspace{-1em}
\end{table}

\section{Conclusion}
We introduce Advantage Decoupled Preference Optimization (ADPO), a reinforcement learning framework that trains a unified policy to both generate solutions and verification scores. ADPO addresses three key challenges in parallel test-time scaling: (i) it enables reliable parallel best-of-N selection through unified generator-verifier training; (ii) it replaces binary verification reward with preference verification reward that improve calibration across both discrete and continuous tasks; and (iii) it employs decoupled advantage to separate gradient flows for generation and verification, thereby mitigating reward hacking and gradient interference. Extensive evaluation across five benchmarks spanning three domains: MathVista, MMMU, ReasonSeg, AndroidControl, and GUI Odyssey, which demonstrates that ADPO achieves superior pass@1 performance while consistently improving best-of-N selection and delivering superior self-verification calibration.

{
    \small
    \bibliographystyle{ieeenat_fullname}
    \bibliography{main}
}
\clearpage
\appendix
\section{Training Details}
The key optimization and training hyperparameters for all experiments are summarized in Table~\ref{tab:hyperparams}.For all experiments, we fully fine-tune with freezing vision modules.
Unless otherwise specified, we conduct training on a single node with 8 NVIDIA A100 GPUs.

For multimodal math reasoning and visual grounding, we prompt the model first to generate its reasoning within \texttt{<think>...</think>}, then produce a final answer in \texttt{<answer>...</answer>}, and finally output a verification score in \texttt{<score>...</score>}.
Reward function consists of accuracy and formatting components. 
The accuracy component uses the widely adopted Python library math-verify to extract the model's answer and compare it with the ground truth. 
The format component ensures the correct structure and ordering of the \texttt{<think>} and \texttt{<answer>} tags, as well as the \texttt{<score>} tag.

For the math reasoning domain, we follow the open-r1-multimodal settings.
We train on the curated multimodal math reasoning dataset, containing curated multimodal math problems with images and answers and evaluated on MathVista and MMMU.
For the visual grounding tasks, we follow the VLM\_R1 settings.
We train on RefCOCO training set and evaluate on ReasonSeg.

For mobile agent tasks, we prompt the model dirctly to generate the tool call following the Qwen2.5-VL mobile use format in \texttt{<tool\_call>...</tool\_call>} and then output the verification score in \texttt{<score>...</score>}.
Each sample contains a resized high-resolution screenshot, a natural-language goal together with a history of previous actions (\texttt{pre\_act}), and a ground-truth tool call of the form \texttt{mobile\_use(...)} specifying the action type (click, swipe, type, open, system\_button, etc.) and its parameters (coordinates, text, button type, and so on).

The answer reward comprises both format and accuracy components. 
The format component ensures the correct structure and ordering of the \texttt{<tool\_call>} and \texttt{<score>} tags. 
The accuracy component follows a stepwise verification process.
First, we validate the action type against the ground truth, granting a $+1$ reward for a match. 
Subsequently, we evaluate the action parameters.
 For example, a click action earns an additional $+1$ reward if the Euclidean distance between the predicted coordinates and the ground truth label is less than $0.14 \times \text{screen diagonal}$.
\begin{table}[h]
    \centering
    \begin{tabular}{ll}
        \hline
        \textbf{Hyperparameter}         & \textbf{Value}      \\
        \hline
        num\_generations                & 8                   \\
        per\_device\_train\_batch\_size & 8                   \\
        gradient\_accumulation\_steps   & 2                   \\
        torch\_dtype                    & bfloat16            \\
        data\_seed                      & 42                  \\
        gradient\_checkpointing         & true                \\
        attn\_implementation            & flash\_attention\_2 \\
        learning\_rate                  & 1e-6                \\
        $\beta$                         & 0.01                \\
        \hline
    \end{tabular}
    \caption{
        Hyperparameter settings used in the training experiments.
    }
    \label{tab:hyperparams}
\end{table}

\section{Ablation and Analysis}

\begin{figure}[t]
    \centering
    \includegraphics[width=\linewidth]{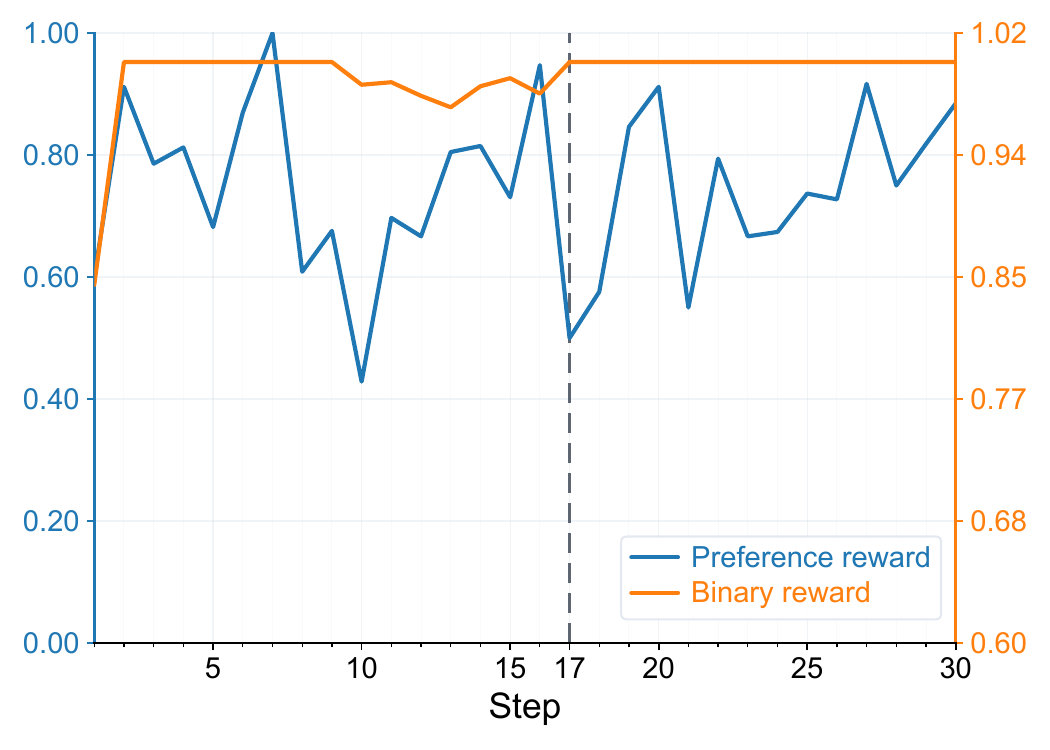}
    \caption{
        \textbf{Analysis of reward signal distribution during training.}
        The \textcolor[rgb]{0.122,0.467,0.706}{\textbf{Blue}} line shows the proportion of correct answers among responses with preference verification reward = 1.
        The \textcolor[rgb]{1.000,0.498,0.055}{\textbf{Orange}} line shows the proportion of correct answers among responses with binary verification reward = 1.
    }
    \vspace{-1.5em}
    \label{fig:appendix_reward_ratio}
\end{figure}

To better understand the effect of our preference verification reward in the mobile agent setting, we analyze training dynamics on the AndroidControl.
As the policy improves, the binary verification reward quickly becomes dominated by correct trajectories, leading to a severe class imbalance where almost all reward-1 samples correspond to already-correct actions.
As shown in \cref{fig:appendix_reward_ratio}, the subset of rollouts that receive binary reward~$=1$ rapidly collapses to near-perfect accuracy, providing little signal to separate moderate-quality actions from the very best ones.
In contrast, the preference verification reward maintains a more informative mixture of correct and incorrect rollouts among its positive signals, preserving contrastive supervision even when overall task success is high.

\begin{figure}[t]
    \centering
    \includegraphics[width=\linewidth]{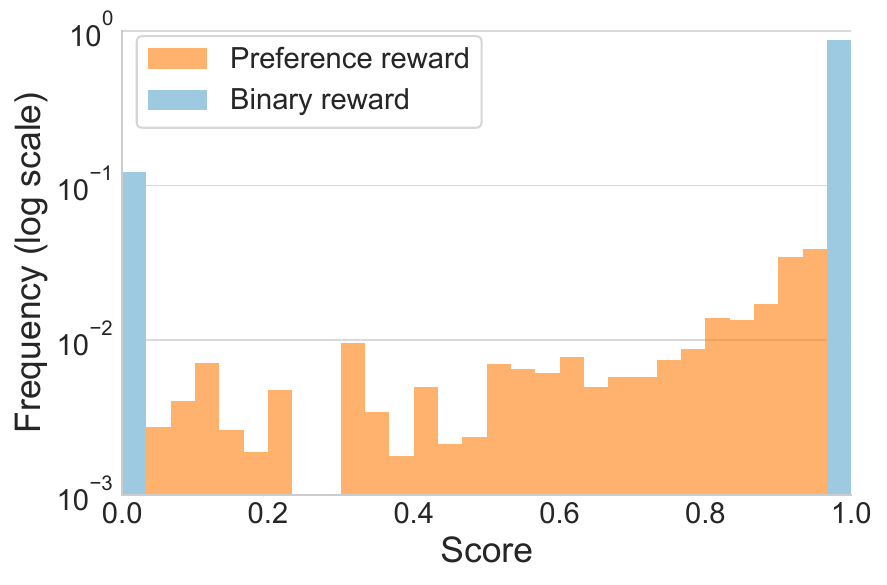}
    \caption{
        \textbf{Comparison of score distributions between models trained with Binary Reward and Preference Reward.}
        While the binary reward model tends to output discrete scores, the preference reward model produces a more diverse distribution. 
    }
    \vspace{-1.5em}
    \label{fig:appendix_score_distribution}
\end{figure}

This difference in supervision is reflected in the learned verification scores.
Figure~\ref{fig:appendix_score_distribution} compares score distributions for models trained with binary versus preference verification reward on AndroidControl.
The binary reward model tends to output highly discretized scores concentrated near the extremes, consistent with the imbalanced 0/1 supervision.
In contrast, the preference reward model produces a smoother and more diverse score distribution, assigning fine-grained scores that better reflect relative action quality.
This diversity is crucial for best-of-$N$ selection, where ranking among multiple candidate trajectories matters more than predicting a single calibrated probability.

Tables~\ref{tab:ablation_reward} and~\ref{tab:ablation_adv} provide more fine-grained ablation results complementing the main figures.
For MathVista and AndroidControl, we report task performance using answer accuracy under pass@1 and best@8, while for ReasonSeg we use $\text{acc@}(\text{IoU}>0.5)$ under the same best-of-$k$ protocol; all three domains share the same self-verification metrics AUC and AP.
In the reward ablation, replacing the binary verification reward with our preference verification reward consistently strengthens self-verification, with especially large gains in AUC and AP on AndroidControl, indicating a much more reliable verifier under class imbalance.
In the advantage ablation, decoupled advantage improves both pass@1 and best@8 performance across domains compared to entangled advantage, while also enhancing verification metrics, showing that separating generation and verification advantages benefits both task accuracy and ranking quality.

\begin{table}[!htbp]
    \centering
    \vspace{-1em}
    \caption{
        \textbf{Ablation of Preference reward.}
        Replacing the binary answer reward with our \emph{preference reward} consistently strengthens self-verification ($\uparrow$AUC/AP) and improves best of N selection performance on \textit{Math}, \textit{Grounding}, and \textit{GUI Agent}.
    }
    \label{tab:ablation_reward}
    \resizebox{\linewidth}{!}{
        \begin{tabular}{ll|cc|cc}
            \toprule
            \multirow{2}{*}{\textbf{Domain}} & \multirow{2}{*}{\textbf{Method}} & \multicolumn{2}{c|}{\textbf{Performance}} & \multicolumn{2}{c}{\textbf{Verification}}                                   \\
                                             &                                  & \textbf{pass@1}                           & \textbf{best@8}                           & \textbf{AUC}   & \textbf{AP}    \\
            \midrule
            \multirow{2}{*}{MathVista}            & Binary verification reward                    & 61.9                                      & 63.4                                      & 0.509          & 0.640          \\
                                             & Preference verification reward         & \textbf{62.4}                             & \textbf{65.0}                             & \textbf{0.522} & \textbf{0.653} \\
            \midrule
            \multirow{2}{*}{ReasonSeg}       & Binary verification reward                    & \textbf{71.8}                             & 72.6                                      & 0.636          & 0.789          \\
                                             & Preference verification reward         & 71.7                                      & \textbf{73.5}                             & \textbf{0.672} & \textbf{0.804} \\
            \midrule
            \multirow{2}{*}{AndroidControl}       & Binary verification reward                    & 70.6                                      & 70.6                                      & 0.609          & 0.765          \\
                                             & Preference verification reward         & \textbf{70.9}                             & \textbf{72.7}                             & \textbf{0.727} & \textbf{0.841} \\
            \bottomrule
        \end{tabular}
    }
\end{table}
\begin{table}[!htbp]
    \centering
    \caption{
        \textbf{Ablation study on decoupled advantages.}
        Our advantage decoupled optimization consistently outperforms entangled advantage in both task performance and solution verification across mathematical reasoning, grounding, and GUI agent tasks.
    }
    \label{tab:ablation_adv}
    \resizebox{\linewidth}{!}{
        \begin{tabular}{ll|cc|cc}
            \toprule
            \multirow{2}{*}{\textbf{Domain}} & \multirow{2}{*}{\textbf{Method}} & \multicolumn{2}{c|}{\textbf{Performance}} & \multicolumn{2}{c}{\textbf{Verification}}                                   \\
                                             &                                  & \textbf{pass@1}                           & \textbf{best@8}                           & \textbf{AUC}   & \textbf{AP}    \\
            \midrule
            \multirow{2}{*}{MathVista}            & Entangled Advantage               & 61.5                                      & 63.1                                      & 50.4           & 61.8           \\
                                             & Decoupled Advantage          & \textbf{62.4}                             & \textbf{65.0}                             & \textbf{52.2}  & \textbf{65.3}  \\
            \midrule
            \multirow{2}{*}{ReasonSeg}       & Entangled Advantage               & 70.4                                      & 71.0                                      & 0.641          & 0.774          \\
                                             & Decoupled Advantage          & \textbf{71.7}                             & \textbf{73.5}                             & \textbf{0.672} & \textbf{0.804} \\
            \midrule
            \multirow{2}{*}{AndroidControl}       & Entangled Advantage               & 71.0                                      & 70.4                                      & 0.542          & 0.715          \\
                                             & Decoupled Advantage          & \textbf{70.9}                             & \textbf{72.7}                             & \textbf{0.727} & \textbf{0.841} \\
            \bottomrule
        \end{tabular}
    }
\end{table}
\section{Extended Experimental Results}

We conduct extensive experiments by enumerating all combinations of base, GRPO, and ADPO as both generators and verifiers across math reasoning, visual grounding, and mobile agent benchmarks (~\cref{tab:math,tab:grounding,tab:agent}).
These results show that ADPO matches GRPO in pass@1 generation quality while providing substantially stronger verification for best-of-$N$ selection, and that this unified training only incurs about 10\% additional training time compared to GRPO with exactly the same data.
In contrast to traditional pipelines that train a separate reward model with extra preference data, ADPO jointly learns generation and verification within a single policy, avoiding additional data collection and separate training runs and thus reducing both training and deployment cost.

Table~\ref{tab:math} reports extended MathVista and MMMU results. With single-sample decoding (Sample 1), GRPO and ADPO generators reach similar pass@1 accuracy (62.2\% vs 62.4\% on MathVista; 48.7\% vs 47.7\% on MMMU), showing ADPO matches GRPO. Under best-of-12 decoding (Sample 12), pairing the ADPO generator and verifier increases MathVista accuracy from 62.5\% to 65.3\% and MMMU from 50.8\% to 52.3\%, indicating a stronger verifier under the same sampling budget.

Table~\ref{tab:grounding} gives analogous ReasonSeg visual grounding results across short, long, and overall queries with gIoU, cIoU, and $\text{acc@}(\text{IoU}>0.5)$. On Sample 1, overall accuracy rises from 68.4\% for the base generator to 71.1\% for GRPO and 71.7\% for ADPO. For best-of-12 (Sample 12), using ADPO for both generation and verification attains 73.2\% overall, outperforming GRPO–GRPO (71.7\%) and the base generator with majority voting (69.1\%), confirming ADPO as the strongest box-ranking verifier.

Table~\ref{tab:agent} summarizes mobile agent results on AndroidControl and GUI Odyssey using step success rate (SR) and related metrics. On Sample 1, GRPO and ADPO generators achieve nearly identical SR (71.0\% vs 70.9\% on AndroidControl; 79.8\% vs 79.7\% on GUI Odyssey), indicating no pass@1 degradation. Under best-of-8 (Sample 8), the ADPO generator–verifier pair improves SR from 70.9\% to 72.7\% on AndroidControl and from 80.6\% to 81.7\% on GUI Odyssey, while the base generator with majority voting lags behind (58.3\% and 46.6\%), quantifying ADPO's stronger verification despite identical training data.

\begin{table*}[!htbp]
    \centering
    \caption{
        \textbf{Evaluation results on multimodal math reasoning benchmarks.}
        Rows correspond to generators and columns correspond to verifiers.
        We use Qwen2-VL-7B as the base model, with GRPO and ADPO representing the finetuned models.
        Majority voting serves as the verifier baseline.
        Values are accuracy (\%).
        GVQA: General VQA; MVQA: Math Target VQA; ARD: Art \& Design; BUS: Business; HEM: Health \& Medicine; HSS: Human \& Social Science; SCI: Science; TEN: Technology \& Engineering.
    }
    \label{tab:math}
    \resizebox{0.7\linewidth}{!}{
        \setlength{\tabcolsep}{2pt}
        \begin{tabular}{cc|ccc|ccccccc}
            \toprule
            \multirow{2}{*}{\textbf{Generator}} & \multirow{2}{*}{\textbf{Verifier}} & \multicolumn{3}{c|}{\textbf{MathVista (In-domain)}} & \multicolumn{7}{c}{\textbf{MMMU (OOD)}}                                                                                                                                 \\
                                                &                                    & GVQA                                                & MVQA                                    & \textbf{ALL}  & ARD           & BUS           & HEM           & HSS           & SCI           & TEN           & \textbf{ALL}  \\
            \midrule
            \rowcolor{backblue2} \multicolumn{12}{c}{\textit{Sample 1}}                                                                                                                                                                                                                                              \\
            \midrule
            Base                                & \xmark                             & 68.9                                                & 48.5                                    & 57.9          & 67.5 & 39.1          & 49.3          & 69.0          & 33.9          & 36.7          & 47.1          \\
            GRPO                                & \xmark                             & 69.8                                       & 55.7                                    & 62.2          & 65.0          & 45.9          & 48.2          & 68.2          & 35.9 & 39.8 & 48.7 \\
            ADPO                                & \xmark                             & 68.7                                                & 57.0                           & 62.4 & 63.1          & 46.2 & 50.2 & 71.1 & 33.3          & 35.3          & 47.7          \\
            \midrule
            \rowcolor{backblue2} \multicolumn{12}{c}{\textit{Sample 4}}                                                                                                                                                                                                                                              \\
            \midrule
            \multirow{5}{*}{Base}               & Major                              & 65.7                                                & 51.9                                    & 58.2          & 66.7          & 47.3          & 50.7          & 65.8          & 34.0          & 38.1          & 48.6          \\
                                                & Base                               & 63.9                                                & 48.7                                    & 55.7          & 60.0          & 44.0          & 50.7          & 60.8          & 32.7          & 33.8          & 45.2          \\
                                                & GRPO                               & 63.3                                                & 48.9                                    & 55.5          & 61.7          & 43.3          & 50.0          & 63.3          & 30.0          & 36.7          & 45.8          \\
                                                & ADPO                               & 63.3                                                & 50.6                                    & 56.4          & 66.7          & 49.3          & 53.3          & 70.8          & 34.0          & 37.6          & 49.9          \\
            \noalign{\vskip 2pt} \cdashline{1-12} \noalign{\vskip 2pt}
            \multirow{5}{*}{GRPO}               & Major                              & 69.8                                                & 58.0                                    & 63.4          & 65.8          & 44.7          & 50.0          & 70.0          & 42.0          & 36.7          & 49.4          \\
                                                & Base                               & 70.2                                                & 55.7                                    & 62.4          & 62.5          & 44.0          & 50.0          & 70.8          & 40.0          & 39.5          & 49.3          \\
                                                & GRPO                               & 69.6                                                & 55.7                                    & 62.1          & 62.5          & 44.0          & 50.0          & 70.8          & 40.0          & 39.5          & 49.9          \\
                                                & ADPO                               & 69.6                                                & 55.6                                    & 62.0          & 64.2          & 46.7          & 50.7          & 71.7          & 39.3          & 39.5          & 50.1          \\
            \noalign{\vskip 2pt} \cdashline{1-12} \noalign{\vskip 2pt}
            \multirow{5}{*}{ADPO}               & Major                              & 71.7                                                & 56.1                                    & 63.3          & 66.7          & 48.0          & 52.7          & 70.0          & 39.3          & 39.0          & 50.7          \\
                                                & Base                               & 68.5                                                & 55.6                                    & 61.5          & 64.2          & 44.0          & 52.7          & 67.5          & 36.7          & 36.7          & 48.3          \\
                                                & GRPO                               & 68.3                                                & 56.9                                    & 62.1          & 65.0          & 44.0          & 52.0          & 68.3          & 40.7          & 36.7          & 49.1          \\
                                                & ADPO                               & 71.3                                                & 59.3                                    & 64.8          & 68.3          & 48.0          & 52.0          & 69.2          & 39.3          & 39.5          & 50.8          \\
            \midrule
            \rowcolor{backblue2} \multicolumn{12}{c}{\textit{Sample 8}}                                                                                                                                                                                                                                              \\
            \midrule
            \multirow{5}{*}{Base}               & Major                              & 68.0                                                & 53.3                                    & 60.1          & 68.3          & 50.0          & 53.3          & 68.3          & 32.7          & 36.7          & 49.4          \\
                                                & Base                               & 63.0                                                & 51.9                                    & 57.0          & 65.0          & 40.7          & 48.0          & 65.0          & 32.0          & 32.4          & 45.0          \\
                                                & GRPO                               & 62.8                                                & 50.9                                    & 56.4          & 65.8          & 40.0          & 49.3          & 65.8          & 32.7          & 37.1          & 46.6          \\
                                                & ADPO                               & 63.5                                                & 50.6                                    & 56.5          & 67.5          & 48.7          & 54.0          & 71.7          & 36.0          & 41.0          & 51.2          \\
            \noalign{\vskip 2pt} \cdashline{1-12} \noalign{\vskip 2pt}
            \multirow{5}{*}{GRPO}               & Major                              & 70.4                                                & 56.5                                    & 62.9          & 66.7          & 48.7          & 51.3          & 74.2          & 42.7          & 36.7          & 51.1          \\
                                                & Base                               & 67.6                                                & 55.0                                    & 60.7          & 62.5          & 47.3          & 51.3          & 72.5          & 38.7          & 37.6          & 49.7          \\
                                                & GRPO                               & 67.6                                                & 55.0                                    & 60.8          & 62.5          & 46.7          & 50.0          & 70.0          & 39.3          & 38.6          & 49.3          \\
                                                & ADPO                               & 67.6                                                & 54.4                                    & 60.5          & 63.3          & 46.7          & 53.3          & 68.3          & 38.7          & 39.0          & 49.8          \\
            \noalign{\vskip 2pt} \cdashline{1-12} \noalign{\vskip 2pt}
            \multirow{5}{*}{ADPO}               & Major                              & 71.1                                                & 58.0                                    & 64.0          & 65.0          & 49.3          & 56.7          & 71.7          & 38.7          & 40.5          & 51.8          \\
                                                & Base                               & 70.0                                                & 55.7                                    & 62.3          & 63.3          & 52.0          & 53.3          & 66.7          & 42.7          & 41.0          & 51.6          \\
                                                & GRPO                               & 69.8                                                & 55.9                                    & 62.3          & 63.3          & 52.7          & 54.7          & 65.8          & 42.0          & 39.0          & 51.2          \\
                                                & ADPO                               & 72.2                                                & 58.9                                    & 65.0          & 65.8          & 54.0          & 54.7          & 66.7          & 40.7          & 41.0          & 52.1          \\
            \midrule
            \rowcolor{backblue2} \multicolumn{12}{c}{\textit{Sample 12}}                                                                                                                                                                                                                                             \\
            \midrule
            \multirow{5}{*}{Base}               & Major                              & 67.4                                                & 55.0                                    & 60.7          & 69.2          & 52.0          & 50.7          & 70.8          & 38.0          & 36.7          & 50.7          \\
                                                & Base                               & 63.7                                                & 51.1                                    & 56.9          & 59.2          & 47.3          & 51.3          & 64.2          & 30.0          & 33.8          & 45.8          \\
                                                & GRPO                               & 62.4                                                & 51.1                                    & 56.3          & 58.3          & 45.3          & 49.3          & 63.3          & 30.0          & 35.2          & 45.2          \\
                                                & ADPO                               & 62.6                                                & 48.5                                    & 55.0          & 65.0          & 52.7          & 53.3          & 70.0          & 40.0          & 35.2          & 50.6          \\
            \noalign{\vskip 2pt} \cdashline{1-12} \noalign{\vskip 2pt}
            \multirow{5}{*}{GRPO}               & Major                              & 70.7                                                & 57.2                                    & 63.4          & 64.2          & 50.0          & 51.3          & 73.3          & 43.3          & 39.5          & 51.7          \\
                                                & Base                               & 70.0                                                & 56.3                                    & 62.6          & 63.3          & 48.0          & 54.0          & 68.3          & 42.7          & 41.0          & 51.2          \\
                                                & GRPO                               & 69.3                                                & 56.7                                    & 62.5          & 63.3          & 48.7          & 52.7          & 67.5          & 42.0          & 40.5          & 50.8          \\
                                                & ADPO                               & 69.6                                                & 55.2                                    & 61.8          & 64.2          & 48.0          & 52.7          & 69.2          & 43.3          & 41.0          & 51.3          \\
            \noalign{\vskip 2pt} \cdashline{1-12} \noalign{\vskip 2pt}
            \multirow{5}{*}{ADPO}               & Major                              & 72.0                                                & 58.3                                    & 64.6          & 65.8          & 50.0          & 53.3          & 75.0          & 36.7          & 39.0          & 51.2          \\
                                                & Base                               & 70.7                                                & 56.5                                    & 63.0          & 62.5          & 52.0          & 54.7          & 70.0          & 41.3          & 41.4          & 52.0          \\
                                                & GRPO                               & 71.3                                                & 56.9                                    & 63.5          & 63.3          & 53.3          & 52.7          & 70.8          & 41.3          & 43.3          & 52.6          \\
                                                & ADPO                               & 71.7                                                & 59.8                                    & 65.3          & 67.5          & 53.3          & 54.0          & 71.7          & 38.7          & 40.5          & 52.3          \\
            \bottomrule
        \end{tabular}
    }
\end{table*}

\begin{table*}[!htbp]
    \centering
    \caption{
        \textbf{Evaluation results on image grounding benchmarks.}
        Rows correspond to generators and columns correspond to verifiers. We use Qwen2.5-VL-7B as the base model, with GRPO and ADPO representing the finetuned models. Majority voting serves as the verifier baseline.
        Models are trained on RefCOCO training set and tested on ReasonSeg (out-of-domain).
    }
    \label{tab:grounding}
    \resizebox{0.8\linewidth}{!}{
        \begin{tabular}{cc|ccc|ccc|ccc}
        \toprule
        \multirow{2}{*}{\textbf{Generator}} & \multirow{2}{*}{\textbf{Verifier}} & \multicolumn{3}{c|}{\textbf{Short query}} & \multicolumn{3}{c|}{\textbf{Long query}} & \multicolumn{3}{c}{\textbf{Overall}}                                                                                                 \\
                                            &                                    & \textbf{gIoU}                             & \textbf{cIoU}                            & \textbf{ACC}                         & \textbf{gIoU} & \textbf{cIoU} & \textbf{ACC}  & \textbf{gIoU} & \textbf{cIoU} & \textbf{ACC}  \\
        \midrule
        \rowcolor{backblue2} \multicolumn{11}{c}{\textit{Sample 1}}                                                                                                                                                                                                                                            \\
        \midrule
        Base                                & \xmark                             & 49.5                                      & 53.0                                     & 67.0                                 & 56.8          & 57.5          & 68.5          & 56.3          & 57.2          & 68.4          \\
        GRPO                                & \xmark                             & 51.8                                      & 55.5                                     & 67.9                                 & 59.1          & 59.7          & 71.3          & 58.6          & 59.5          & 71.1          \\
        ADPO                                & \xmark                             & 51.7                                      & 54.8                                     & 68.0                        & 60.2          & 59.4          & 71.9 & 58.1          & 59.1          & 71.7 \\
        \midrule
        \rowcolor{backblue2} \multicolumn{11}{c}{\textit{Sample 4}}                                                                                                                                                                                                                                            \\
        \midrule
        \multirow{5}{*}{Base}               & Major                              & 47.8                                      & 52.0                                     & 66.0                                 & 57.3          & 57.9          & 69.3 & 56.7          & 57.5          & 69.1 \\\
                                            & Base                               & 47.2                                      & 51.4                                     & 66.0                                 & 56.9          & 57.5          & 68.6          & 56.3          & 57.1          & 68.4          \\
                                            & GRPO                               & 49.6                                      & 53.2                                     & 67.0                                 & 57.4          & 57.9          & 68.7          & 56.9          & 57.7          & 68.6          \\
                                            & ADPO                               & 50.4                                      & 53.5                                     & 68.0                        & 57.3          & 57.9          & 69.1          & 56.9          & 57.7          & 69.1 \\
        \noalign{\vskip 2pt} \cdashline{1-11} \noalign{\vskip 2pt}
        \multirow{5}{*}{GRPO}               & Major                              & 54.5                                      & 57.0                                     & 68.0                                 & 58.8          & 59.5          & 72.1          & 58.5          & 59.4          & 71.8          \\
                                            & Base                               & 55.2                                      & 57.5                                     & 68.0                                 & 59.7          & 60.3          & 72.9          & 59.4          & 60.2          & 72.6          \\
                                            & GRPO                               & 53.4                                      & 56.4                                     & 68.9                        & 59.1          & 59.7          & 72.0          & 58.8          & 59.5          & 71.8          \\\
                                            & ADPO                               & 55.1                                      & 57.8                                     & 68.0                                 & 60.5          & 61.1          & 73.4 & 60.2          & 60.9          & 73.1 \\
        \noalign{\vskip 2pt} \cdashline{1-11} \noalign{\vskip 2pt}
        \multirow{5}{*}{ADPO}               & Major                              & 52.7                                      & 55.3                            & 67.0                                 & 60.1          & 60.5          & 72.0          & 59.6          & 60.2          & 71.7          \\\
                                            & Base                               & 51.2                                      & 54.1                                     & 66.0                                 & 59.3          & 59.9          & 71.7          & 58.8          & 59.6          & 71.4          \\
                                            & GRPO                               & 51.0                                      & 54.3                                     & 67.0                                 & 60.0          & 60.7          & 72.7          & 59.4          & 60.3          & 72.4          \\
                                            & ADPO                               & 52.2                                      & 55.1                                     & 67.0                                 & 61.0          & 61.5          & 73.3 & 60.5          & 61.1          & 72.9 \\
        \midrule
        \rowcolor{backblue2} \multicolumn{11}{c}{\textit{Sample 8}}                                                                                                                                                                                                                                            \\
        \midrule
        \multirow{5}{*}{Base}               & Major                              & 47.8                                      & 51.4                                     & 63.1                        & 57.2          & 57.8          & 69.2          & 56.6          & 57.4          & 68.8          \\\
                                            & Base                               & 47.9                                      & 51.4                                     & 62.1                                 & 56.6          & 57.2          & 68.1          & 56.1          & 56.9          & 67.7          \\
                                            & GRPO                               & 47.1                                      & 50.5                                     & 62.1                                 & 56.8          & 57.5          & 68.4          & 56.2          & 57.0          & 68.0          \\
                                            & ADPO                               & 47.4                                      & 50.9                                     & 61.2                                 & 57.7          & 58.4          & 69.4 & 57.1          & 57.9          & 68.9 \\
        \noalign{\vskip 2pt} \cdashline{1-11} \noalign{\vskip 2pt}
        \multirow{5}{*}{GRPO}               & Major                              & 52.0                                      & 55.6                                     & 68.0                        & 59.2          & 59.9          & 72.0          & 58.7          & 59.6          & 71.7          \\
                                            & Base                               & 51.7                                      & 55.0                                     & 67.0                                 & 60.1          & 60.7          & 72.4          & 59.6          & 60.4          & 72.1          \\
                                            & GRPO                               & 51.5                                      & 55.3                                     & 68.0                        & 60.0          & 60.7          & 72.4          & 59.5          & 60.4          & 72.1          \\
                                            & ADPO                               & 54.4                                      & 57.2                                     & 67.0                                 & 60.6          & 61.3          & 73.8 & 60.2          & 61.1          & 73.3 \\
        \noalign{\vskip 2pt} \cdashline{1-11} \noalign{\vskip 2pt}
        \multirow{5}{*}{ADPO}               & Major                              & 53.2                                      & 56.1                                     & 67.0                                 & 58.8          & 59.4          & 71.3          & 58.5          & 59.2          & 71.0          \\
                                            & Base                               & 52.9                                      & 55.4                                     & 67.0                                 & 59.6          & 60.2          & 72.0          & 59.2          & 59.9          & 71.7          \\
                                            & GRPO                               & 55.6                                      & 57.8                                     & 68.9                        & 59.9          & 60.6          & 72.9          & 59.6          & 60.5          & 72.7          \\\
                                            & ADPO                               & 53.2                                      & 56.0                                     & 67.0                                 & 60.9          & 61.5          & 73.7 & 60.4          & 61.2          & 73.5 \\
        \midrule
        \rowcolor{backblue2} \multicolumn{11}{c}{\textit{Sample 12}}                                                                                                                                                                                                                                           \\
        \midrule
        \multirow{5}{*}{Base}               & Major                              & 50.2                                      & 53.7                                     & 66.0                                 & 57.2          & 57.8          & 69.3 & 56.8          & 57.6          & 69.1 \\\
                                            & Base                               & 49.5                                      & 53.3                                     & 68.0                        & 56.9          & 57.6          & 68.9          & 56.5          & 57.4          & 68.8          \\
                                            & GRPO                               & 50.0                                      & 53.1                                     & 66.0                                 & 57.1          & 57.9          & 69.1          & 56.7          & 57.6          & 68.9          \\
                                            & ADPO                               & 49.8                                      & 52.6                                     & 65.1                                 & 57.6          & 58.2          & 69.2          & 57.1          & 57.8          & 68.9          \\
        \noalign{\vskip 2pt} \cdashline{1-11} \noalign{\vskip 2pt}
        \multirow{5}{*}{GRPO}               & Major                              & 55.6                                      & 58.1                                     & 69.9                        & 58.8          & 59.5          & 72.2          & 58.6          & 59.4          & 72.0          \\\
                                            & Base                               & 48.8                                      & 52.4                                     & 65.1                                 & 59.6          & 60.2          & 72.5          & 58.9          & 59.7          & 72.1          \\
                                            & GRPO                               & 53.2                                      & 55.8                                     & 68.0                                 & 59.1          & 59.9          & 71.9          & 58.8          & 59.6          & 71.7          \\
                                            & ADPO                               & 54.1                                      & 56.7                                     & 68.9                                 & 60.9          & 61.6          & 74.0 & 60.5          & 61.3          & 73.7 \\
        \noalign{\vskip 2pt} \cdashline{1-11} \noalign{\vskip 2pt}
        \multirow{5}{*}{ADPO}               & Major                              & 53.3                                      & 55.3                                     & 66.0                                 & 59.3          & 60.0          & 71.8          & 58.9          & 59.8          & 71.5          \\
                                            & Base                               & 55.0                                      & 57.4                                     & 68.9                        & 60.2          & 60.9          & 72.1          & 59.9          & 60.7          & 71.9          \\\
                                            & GRPO                               & 53.8                                      & 56.5                                     & 68.9                                 & 60.3          & 61.0          & 72.4          & 59.9          & 60.7          & 72.1          \\
                                            & ADPO                               & 53.9                                      & 56.2                                     & 67.0                                 & 61.3          & 62.0          & 73.6 & 60.9          & 61.6          & 73.2 \\
        \bottomrule
    \end{tabular}
    }
\end{table*}

\begin{table*}[!htbp]
    \centering
    \caption{
        \textbf{Evaluation results on mobile agent benchmarks.}
        Rows correspond to generators and columns correspond to verifiers.
        We use Qwen2.5-VL-7B as the base model, with GRPO and ADPO representing the finetuned models.
    }
    \label{tab:agent}
    \resizebox{0.7\linewidth}{!}{
        \begin{tabular}{cc|ccc|ccc}
            \toprule
            \multirow{2}{*}{\textbf{Generator}} & \multirow{2}{*}{\textbf{Verifier}} & \multicolumn{3}{c|}{\textbf{AndroidControl}} & \multicolumn{3}{c}{\textbf{GUI Odyssey}}                                                                 \\
                                                &                                    & Type                                         & Grounding                                & \textbf{SR}   & Type          & Grounding     & \textbf{SR}   \\
            \midrule
            \rowcolor{backblue2} \multicolumn{8}{c}{\textit{Sample 1}}                                                                                                                                                                         \\
            \midrule
            Base                                & \xmark                             & 82.2                                         & 73.6                                     & 61.3          & 81.1          & 61.4          & 52.8          \\
            GRPO                                & \xmark                             & 86.0                                & 76.9                            & 71.0 & 93.1          & 83.9 & 79.8 \\
            ADPO                                & \xmark                             & 85.8                                         & 76.2                                     & 70.9          & 94.2 & 82.5          & 79.7          \\
            \midrule
            \rowcolor{backblue2} \multicolumn{8}{c}{\textit{Sample 4}}                                                                                                                                                                         \\
            \midrule
            \multirow{4}{*}{Base}               & Major                              & 76.3                                         & 68.1                                     & 56.0          & 76.9          & 55.3          & 46.5          \\
                                                & Base                               & 72.1                                         & 67.8                                     & 52.5          & 75.3          & 55.6          & 45.2          \\
                                                & GRPO                               & 74.9                                         & 71.4                                     & 57.7          & 75.3          & 55.2          & 45.3          \\
                                                & ADPO                               & 76.4                                         & 74.5                                     & 60.7          & 75.1          & 55.7          & 45.6          \\
            \noalign{\vskip 2pt} \cdashline{1-8} \noalign{\vskip 2pt}
            \multirow{4}{*}{GRPO}               & Major                              & 85.5                                         & 77.2                                     & 71.0          & 94.7          & 83.9          & 81.3          \\
                                                & Base                               & 85.4                                         & 77.2                                     & 71.0          & 94.3          & 83.8          & 81.0          \\
                                                & GRPO                               & 85.4                                         & 77.3                                     & 70.8          & 94.4          & 83.7          & 80.7          \\
                                                & ADPO                               & 85.6                                         & 77.7                                     & 71.2          & 94.5          & 84.0          & 81.4          \\
            \noalign{\vskip 2pt} \cdashline{1-8} \noalign{\vskip 2pt}
            \multirow{4}{*}{ADPO}               & Major                              & 86.6                                         & 77.1                                     & 71.6          & 93.9          & 83.5          & 79.8          \\
                                                & Base                               & 86.4                                         & 76.4                                     & 71.0          & 94.7          & 84.2          & 81.2          \\
                                                & GRPO                               & 86.4                                         & 77.9                                     & 72.0          & 94.7          & 84.2          & 81.1          \\
                                                & ADPO                               & 86.3                                         & 79.5                                     & 72.7          & 94.7          & 84.5          & 81.6          \\
            \midrule
            \rowcolor{backblue2} \multicolumn{8}{c}{\textit{Sample 8}}                                                                                                                                                                         \\
            \midrule
            \multirow{4}{*}{Base}               & Major                              & 78.7                                         & 68.8                                     & 58.3          & 76.7          & 55.4          & 46.6          \\
                                                & Base                               & 73.9                                         & 68.4                                     & 54.3          & 75.1          & 55.2          & 44.9          \\
                                                & GRPO                               & 77.3                                         & 73.4                                     & 61.0          & 74.4          & 54.3          & 44.5          \\
                                                & ADPO                               & 79.7                                         & 76.5                                     & 64.7          & 73.9          & 54.6          & 44.6          \\
            \noalign{\vskip 2pt} \cdashline{1-8} \noalign{\vskip 2pt}
            \multirow{4}{*}{GRPO}               & Major                              & 85.6                                         & 76.9                                     & 70.8          & 94.6          & 84.4          & 81.5          \\
                                                & Base                               & 85.6                                         & 77.1                                     & 71.0          & 93.7          & 84.4          & 80.7          \\
                                                & GRPO                               & 85.4                                         & 77.4                                     & 70.9          & 93.7          & 84.4          & 80.6          \\
                                                & ADPO                               & 85.6                                         & 77.7                                     & 71.4          & 93.9          & 84.8          & 81.2          \\
            \noalign{\vskip 2pt} \cdashline{1-8} \noalign{\vskip 2pt}
            \multirow{4}{*}{ADPO}               & Major                              & 86.5                                         & 76.4                                     & 71.3          & 94.8          & 84.0          & 80.9          \\
                                                & Base                               & 86.1                                         & 76.2                                     & 70.8          & 95.1          & 84.6          & 81.6          \\
                                                & GRPO                               & 85.8                                         & 77.7                                     & 71.4          & 94.9          & 84.4          & 81.4          \\
                                                & ADPO                               & 86.4                                & 78.7                            & 72.7 & 94.8 & 84.7 & 81.7 \\
            \midrule
            \rowcolor{backblue2} \multicolumn{8}{c}{\textit{Sample 12}}                                                                                                                                                                        \\
            \midrule
            \multirow{4}{*}{Base}               & Major                              & 78.9                                         & 68.7                                     & 58.3          & 76.9          & 55.5          & 46.9          \\
                                                & Base                               & 73.4                                         & 67.9                                     & 53.6          & 74.5          & 55.5          & 44.6          \\
                                                & GRPO                               & 76.8                                         & 73.2                                     & 60.7          & 73.5          & 54.3          & 44.0          \\
                                                & ADPO                               & 79.2                                         & 76.7                                     & 64.5          & 72.9          & 53.8          & 43.6          \\
            \noalign{\vskip 2pt} \cdashline{1-8} \noalign{\vskip 2pt}
            \multirow{4}{*}{GRPO}               & Major                              & 85.6                                         & 77.4                                     & 71.1          & 94.5          & 84.0          & 81.1          \\
                                                & Base                               & 85.6                                         & 78.0                                     & 71.4          & 93.0          & 84.0          & 79.9          \\
                                                & GRPO                               & 85.4                                         & 77.5                                     & 70.9          & 93.1          & 83.9          & 79.7          \\
                                                & ADPO                               & 85.7                                         & 77.9                                     & 71.5          & 93.2          & 84.2          & 80.3          \\
            \noalign{\vskip 2pt} \cdashline{1-8} \noalign{\vskip 2pt}
            \multirow{4}{*}{ADPO}               & Major                              & 86.6                                         & 76.7                                     & 71.9          & 94.4          & 83.7          & 80.5          \\
                                                & Base                               & 86.5                                         & 76.3                                     & 71.6          & 94.8          & 84.6          & 81.5          \\
                                                & GRPO                               & 85.4                                         & 78.6                                     & 71.9          & 94.6          & 84.1          & 81.1          \\
                                                & ADPO                               & 86.3                                         & 78.9                                     & 72.9          & 94.4          & 84.5          & 81.4          \\
            \bottomrule
        \end{tabular}
    }
\end{table*}

\end{document}